\def\doctype{3}
\documentclass[3p]{elsarticle}

\usepackage{lineno}
\modulolinenumbers[5]
\usepackage{microtype}
\usepackage{amsmath,amssymb,amsfonts,amsthm,thmtools}
\usepackage{textcomp}
\usepackage{graphicx}
\usepackage{subfig}
\usepackage{tipa}
\usepackage{mathtools}
\usepackage{fancyref}
\usepackage{float}
\restylefloat{figure}
\usepackage{mathtools}
\usepackage{fancyhdr}
\usepackage{kantlipsum}
\usepackage{eso-pic}
\usepackage{nomencl}
\usepackage{etoolbox}
\usepackage{dsfont}
\usepackage{eucal}
\usepackage{nccmath}
\usepackage{tikz}
\tikzset{>=latex}
\usetikzlibrary{fit,automata,positioning,angles,quotes}
\usepackage[algo2e,linesnumbered,ruled,lined]{algorithm2e}
\usepackage[toc,page]{appendix}
\usepackage{hyperref}
\hypersetup{
	colorlinks=true,
	linkcolor=black,
	filecolor=black,      
	urlcolor=black,
	citecolor=black,}
\usepackage{lmodern}
\usepackage{multirow}
\usepackage{diagbox}

\newtheorem{assumption}{Assumption}
\newtheorem{proposition}{Proposition}
\newtheorem{corollary}{Corollary}

\newtheorem{thm}{Theorem}
\newtheorem{problem_1}{Problem}
\newcommand{\hos}{\textcolor{black}}

\newcommand{\rbtl}{\textcolor{black}}
\newcommand{\green}{\textcolor{black}}
\newcommand{\aij}{\textcolor{black}}
\declaretheoremstyle[notefont=\bfseries,qed={$\lrcorner$}]{defwithcornerstyle}
\declaretheorem[style=defwithcornerstyle,title=Definition]{definition}

\declaretheoremstyle[bodyfont=\itshape,notefont=\bfseries,qed={$\lrcorner$}]{remarkwithcornerstyle}
\declaretheorem[style=remarkwithcornerstyle,title=Remark]{remark}

\journal{Journal of Artificial Intelligence}

\begin{document}

\begin{frontmatter}
\title{Certified Reinforcement Learning with Logic Guidance\tnoteref{t1,t2}}

\tnotetext[t1]{\textbf{This paper is part of the Special Issue: “Risk-aware Autonomous Systems: Theory and Practice”}.}
\tnotetext[t2]{This work is in part supported by the HiClass project (113213), a partnership between the Aerospace Technology Institute (ATI), Department for Business, Energy and Industrial Strategy (BEIS) and Innovate UK}

\author[msr]{Hosein Hasanbeig \corref{atoxford}}

\author[amazon]{Daniel Kroening \corref{amazonR}}

\cortext[atoxford]{The work reported in this paper was done at Department of Computer Science, University of Oxford, UK.} 

\cortext[amazonR]{The work in this paper was done prior to joining Amazon.}

\author[oxford]{Alessandro Abate}

\ead{alessandro.abate@cs.ox.ac.uk}

\address[msr]{Microsoft Research, United States}
\address[amazon]{Magdalen College, University of Oxford, and Amazon, Inc., United States} 
\address[oxford]{Department of Computer Science, University of Oxford, Parks Rd, OX1 3QD, United Kingdom}

\begin{abstract}
Reinforcement Learning (RL) is a widely employed machine learning
architecture that has been applied to a variety of control problems. 
However, applications in safety-critical domains require a systematic and
formal approach to specifying requirements as tasks or goals.  We propose a
model-free RL algorithm that enables the use of Linear Temporal Logic (LTL)
to formulate a goal for unknown continuous-state/action Markov Decision
Processes (MDPs).  The given LTL property is translated into a
Limit-Deterministic Generalised B\"uchi Automaton (LDGBA), which is then
used to shape a synchronous reward function on-the-fly.  Under certain
assumptions, the algorithm is guaranteed to synthesise a control policy
whose traces satisfy the LTL specification with maximal probability.
\end{abstract}


\end{frontmatter}

\section{Introduction}
\label{sec:intro}

Reinforcement Learning (RL) is an area of machine learning, where an
agent is trained to maximise a reward that is calculated by a user-provided
function.  Key to success in RL is the ability to predict the effect of
picking a particular candidate action on the ultimate reward, and neural
networks, owing to their ability to generalise, have enabled the
application of RL in a broad range of application domains.

A significant barrier to successful application of RL is the setup of the
reward function when requirements (the user's goals or the task that is to
be done) are complex~\cite{garcia}: reward engineering often requires
tedious parameter tuning to map complex goals to an appropriate reward
structure~\cite{reduction}.  As a consequence, the trained agent can be
brittle and the policy it implements can be difficult to interpret.  This is
particularly problematic in safety-critical applications, say when the agent
operates in the proximity of humans.  This gives rise to the need for
provably-correct reward shaping.

\begin{figure}[!t]
	\centering
	\includegraphics[width=0.8\columnwidth]{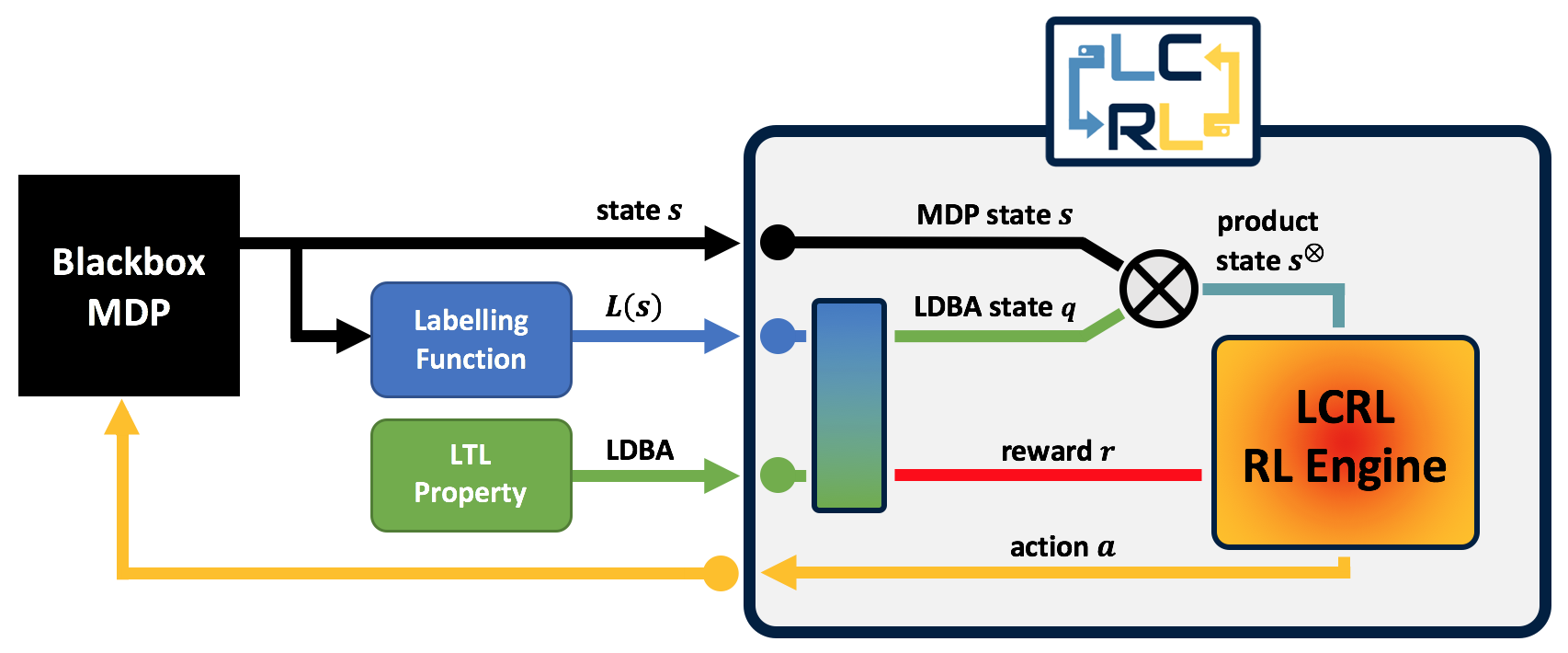}
	\caption{Learning under LTL with Logically-Constrained RL (LCRL). The reward signal is produced automatically by synchronising the LTL property and the unknown MDP.   
		\label{fig:overview}}
\end{figure}

We suggest the use of Linear Temporal Logic (LTL)~\cite{pnueli} as a formal
high-level language to specify complex tasks when applying RL~\cite{clarke}
to sequential decision-making problems.  LTL is a formal language that can
express engineering requirements and specifications, and there exists
a substantial body of research on how to derive LTL properties from natural
language~\cite{natural2LTL,natural2LTL2,natural2LTL3,xinyu2023lang2ltl}.  LTL can express
time-dependent properties, such as safety and liveness, and further allows
to specify tasks with complex dynamics (e.g.,~tasks that are repeating,
sequential, or conditional).

We present an algorithm that, given an LTL formula that describes the goal
the user wants to achieve, automatically shapes a reward function for RL in
a provably-correct manner, and synthesises controllers for which
satisfaction of the given property is guaranteed.  

The standard means for formalising sequential decision-making problems are
Markov Decision Processes (MDPs), a family of stochastic
processes~\cite{puterman}.  In an MDP, a decision maker (or an agent) can
transition between states by choosing relevant actions~\cite{otterlo} while
receiving a scalar reward.  The outcomes of taking actions are, in general,
probabilistic and not fully under the control of the agent~\cite{NDP}.  A
decision-making problem given as an MDP is said to be solved when in any
given state the agent is able to choose the most favourable action so that
the accrued reward is expected to be maximal in the long run~\cite{otterlo}.

When both state and action spaces are finite, the stochastic behaviour of an
MDP can be described by a transition probability matrix.  In this case, MDPs
can be solved via Dynamic Programming (DP).  DP iteratively applies a
Bellman operation on a value function expressing the expected reward of
interest for a state of the MDP~\cite{bellman1957markovian, NDP}.

We can understand convergence properties of RL by relating them to the
optimal value function produced by DP~\cite{watkins,
jaakkola1994convergence}.  When the state and action spaces are not finite,
approximate DP is often employed.  This approximation can be achieved by
generating an abstract model of the MDP itself~\cite{faust-2, faust-1,
faust0, faust}, or by inferring a (non-linear) regression model over the
value function~\cite{cdp-1, cdp0, cdp}.

In practice, however, it is often infeasible to obtain sufficient knowledge of
the stochastic dynamics of the problem, which means that we cannot formulate
the MDP.  In contrast to DP, RL solely depends on a set of experience
samples gathered by exploration, which simplifies the Bellman's backup, and
at the same time avoids exhaustive sweeps over the full state space.  In
this paper, we focus on approaches that are \emph{model-free}, i.e., we do
not require that any model (in the form of an MDP or otherwise) is given. 
Model-free RL methods are very easy to apply and resource-efficient, as the
agent learns an optimal policy without an intermediate model.  Model-free RL
has been shown to converge to the same action selection policy as DP under
mild assumptions~\cite{NDP, watkins}.

Deep RL is arguably one of the most significant breakthroughs in RL, whereby
human-level play has been achieved on a number of Atari games~\cite{deepql}
by incorporating deep neural networks into RL.  This resulted in
successfully tackling StarCraft~\cite{alpha} and the game of Go~\cite{go}. 
Deep-RL-based algorithms are often general enough so that the rules of
different games do not have to be explicitly encoded for the agents to learn
successful control policies.  The success of deep RL has resulted in extensive use of RL,
beyond small-scale environments~\cite{go,alpha,drnn,dqn_robotics}.  In
particular, employment of RL in safety-critical problems has recently been
investigated~\cite{risk1,risk2,risk3,risk4,arxiv,shield,shield2,lcrl_j,carr2022safe},
including autonomous driving~\cite{au_car_1,au_car_2} and
avionics~\cite{initial2,ng}.  This however inevitably entails the need for
correct-by-design policy synthesis, in order to guarantee for instance,
among other quantitative requirements, the safety of policies synthesised
via RL.

\textbf{Contributions:}
We present an algorithm for automatically engineering a reward function
for RL from a given LTL property, and for synthesising a corresponding
optimal policy maximising the probability of satisfying the given LTL
property. Our method is model-free, and allows us to synthesise control
policies under LTL for a continuous-state/action MDP (which subsumes the
simpler, finite state/action special case).  We discuss the assumptions
under which our RL setup is guaranteed to synthesise control policies whose
traces satisfy the LTL specification with maximal probability.

The LTL property offers means to introduce a-priori knowledge of the
structure of the problem into the learning procedure, while avoiding
overfitting demonstrations done by a human teacher.  An LTL property is a
formal, un-grounded, and symbolic representation of the task and of its
steps.  This enables the use of complex properties and, as we show later,
LTL is suitable for sub-task decomposition and hierarchical learning.

In existing methods, the temporal dependency of rewards is often tackled
with \emph{options}~\cite{sutton}, or, in general, dependencies are
structured hierarchically.  Current approaches to hierarchical RL very much
depend on particular state representations and whether they are structured
enough for a suitable reward signal to be effectively engineered manually. 
Hierarchical RL therefore often requires detailed supervision in the form of
explicitly specified high-level actions or intermediate supervisory
signals~\cite{precup, options_1, options_h_1, options_2, options_h_2,
pol-sketch, bacon2017option}.  By contrast, when expressing a complex task
using LTL, each component of the LTL property systematically structures the
complex mission task into low-level, achievable task ``modules" without
requiring any supervision.  The LTL property essentially acts as a
high-level unsupervised guide for the agent, whereas the low-level planning
is handled by a (deep) RL architecture.  To demonstrate this benefit, we
showcase our approach on the Atari 2600 game ``Montezuma's Revenge'', which
is known as an exceptionally hard problem for RL.

Full instructions on reproducing all the case studies in this paper and how to import LCRL into any Python project are provided on the LCRL GitHub page:
\begin{center}
	\url{www.github.com/grockious/lcrl} \texttt{(@8b4e474)}
\end{center}

\section{Overview}

We give a brief overview of our approach.
As is standard in existing methods, we convert the LTL property into an
automaton, namely a finite-state machine~\cite{bible}.  However,
LTL-to-automaton translation may result in a non-deterministic model, over
which policy synthesis for MDPs is not semantically meaningful.  A~standard
solution to this issue is to use Safra's construction to determinise the
automaton.  This is known to increase its size dramatically~\cite{safra,
nba2d}.  An alternative solution is to convert the given LTL formula
directly into a Deterministic Rabin Automaton (DRA).  Such a conversion
results, in the worst case, in automata that are doubly exponential in the
size of the original LTL formula~\cite{dra4}.

By contrast, we propose to express the given LTL property as a
Limit-Deterministic Generalised B\"uchi Automaton (LDGBA)~\cite{sickert}. 
This construction results in a (singly) exponential-sized automaton for
LTL$\setminus$GU\footnote{LTL$\setminus$GU is a fragment of linear temporal
logic with the restriction that no until operator occurs in the scope of an
always operator.} and the result is nearly the same size as a DRA for the
rest of LTL.  We show that this succinctness improves the convergence speed
and sample efficiency of the learning algorithm significantly.  However, the
translation of LTL into LDGBAs introduces non-trivial problems into the
learning algorithm,
which we address in this work.  An additional benefit of our approach is
that LDGBAs are semantically easier than DRAs owing to their acceptance
conditions, which makes policy synthesis algorithms much simpler to
implement~\cite{sickert2, tkachev}.  We emphasise that there exist a variety
of algorithms for constructing limit-deterministic B\"uchi automata from
LTL, but not all of resulting automata can be employed for quantitative
model checking and probabilistic certification~\cite{kini}.  More
importantly, the unique succinctness of the LDGBA
construction~\cite{sickert} used in our work is due to its ``generalised
B\"uchi'' accepting condition, and other, non-generalised constructions
inevitably result in larger automata.

Once the LDGBA is generated from the given LTL property, we construct
on-the-fly\footnote{``On-the-fly'' means that the algorithm tracks (or
executes) the state of an underlying structure (or a function) without
explicitly building the entire structure.} a product between the MDP and the
resulting LDGBA.  The generalised B\"uchi accepting condition then gives
rise to a reward function that is synchronous with the state-action pairs of
the MDP.  Using this algorithmic reward-shaping procedure, an RL scheme is
able to learn an optimal policy that returns the maximum expected reward,
and that equivalently satisfies the given LTL specification with maximal
probability.

Finally, the application of the LDGBA also allows the RL agent to generate
episodes that are informative, particularly for the case of non-Markovian
tasks: we show that this can reduce the sample complexity of RL
architectures.

\section{Background}\label{background}

We consider a universal RL setup, consisting of an agent
interacting with an unknown environment, modelled as a general MDP.

\subsection{Markov Decision Processes}

\begin{definition}[General MDP~\cite{shreve,stochastic}]\label{def:general_mdp} 
The tuple $\mathfrak{M}=\allowbreak(\mathcal{S}, \allowbreak \mathcal{A},
\allowbreak s_0,\allowbreak P,\allowbreak \mathcal{AP},\allowbreak L)$ is a
general MDP over a set of continuous states $\mathcal{S}=\mathbb{R}^n$,
where $\mathcal{A}=\mathbb{R}^{n'}$ is a set of continuous actions, and \rbtl{$\mathcal{S}_0
\subset \mathcal{S}$ is the MDP initial set of states. An initial state $s_0$ is then randomly chosen from $\mathcal{S}_0$}. 
$P:\mathcal{B}(\mathcal{S})\times\mathcal{S}\times\mathcal{A}\rightarrow
[0,1]$ is a Borel-measurable conditional transition kernel which assigns to
any pair of state $s \in \mathcal{S}$ and action $a \in \mathcal{A}$ a
probability measure $P(\cdot|s,a)$ on the Borel space
$(\mathcal{S},\mathcal{B}(\mathcal{S}))$, where $\mathcal{B}(\mathcal{S})$
is the set of all Borel sets on $\mathcal{S}$.  $\mathcal{AP}$ is a finite
set of atomic propositions and a labelling function $L: \mathcal{S}
\rightarrow 2^{\mathcal{AP}}$ assigns to each state $s \in \mathcal{S}$ a
set of atomic propositions $L(s) \subseteq 2^\mathcal{AP}$.
\end{definition}

A finite-state/action MDP is a special case of general MDPs in which
$|\mathcal{S}|<\infty$, $|\mathcal{A}|<\infty$, and
$P:\mathcal{S}\times\mathcal{A}\times\mathcal{S}\rightarrow[0,1]$ is the
transition probability function.  The transition function $P$ reduces to a
transition probability matrix.

A variable $R(s,a)\sim\varUpsilon(\cdot|s,a)\in\mathcal{P}(\mathbb{R^+})$
can be defined over the MDP~$\mathfrak{M}$, representing the immediate
reward obtained when action $a$ is taken at a given state~$s$, where
$\mathcal{P}(\mathbb{R^+})$ is the set of probability distributions on
subsets of $\mathbb{R^+}$, and $\varUpsilon$ is the reward distribution.  A
realisation of $R$ at time step $n$ is denoted as $r_n$.

\begin{definition}[Stationary Deterministic Policy]
A stationary (randomised) policy $\pi: \mathcal{S} \times \mathcal{A}
\rightarrow [0,1]$ is a mapping from any state $s \in \mathcal{S}$ to a
probability distribution over actions.  \hos{A policy $\pi$ assigns to any
state $s\in\mathcal{S}$ a probability measure $\pi(\cdot | s)$ on the Borel
space $(\mathcal{A},\mathcal{B}(\mathcal{A}))$.} A~deterministic policy is a
degenerate case of a randomised policy which outputs a single action at a
given state, that is $\forall s \in \mathcal{S},~\exists a \in
\mathcal{A},~\pi(s,a)=1$.
\end{definition}

\begin{definition}[Expected Infinite-Horizon Discounted Return]
\label{def:expected_utility} 
For any policy $\pi$ on an MDP $\mathfrak{M}$, and given a reward function
$R$, the expected discounted reward return at state $s$ is defined
as~\cite{sutton}:
\begin{align}\label{upol}
\begin{aligned}
&{U}^\pi_\mathfrak{M}(s)=\mathds{E}^\pi[\sum\limits_{n=0}^{\infty} \gamma^n~ r_n|s_0=s], 
&s_n \sim P(\cdot|s_{n-1},a_{n-1}),~a_n \sim \pi(\cdot|s_n), 
\end{aligned}
\end{align}
where $\mathds{E}^{\pi} [\cdot]$ denotes the expected value given that the
agent follows policy $\pi$ from state $ s $, $\gamma\in [0,1)$ ($\gamma\in
[0,1]$ when episodic\footnote{An \emph{episodic} RL algorithm consists of
several reset trajectories, at each of which the agent re-starts from the
initial state of the MDP.}) is a discount factor, and $s_0,a_0,s_1,a_1...$ is
the sequence of states generated by policy $\pi$, initialised at $s_0 = s$. 
We will drop the subscript $\mathfrak{M}$ when it is clear from the context.
\end{definition}

Note that the discount factor $\gamma$ is a hyper-parameter that has to be
tuned.  In particular, there is standard work in RL with state-dependent
discount factors that are shown to preserve convergence and optimality
guarantees, by means of ensuring the contractivity of corresponding
operators required to update value functions or
policies~\cite{discount,discount2,discount3,discount4}.  A~possible tuning
strategy for our setup would allow the discount factor be a function of the
state:
\begin{equation}\label{gamma}
\gamma(s) = \left\{
\begin{array}{ll}
\eta & $ if $ R(s,a)>0,\\
1 & $ otherwise, 
$
\end{array}
\right.
\end{equation}
where $0<\eta<1$ is a constant. Hence, \eqref{upol} would reduce to
\begin{equation}\label{state_dep_utility}
{U}^{\pi}(s)=\mathds{E}^{\pi} [\sum\limits_{n=0}^{\infty} \gamma(s_n)^{N(s_n)}~ r_n)|s_0=s],
\end{equation}
where $N(s_n)$ is the number of times a positive reward has been observed at
state $s_n$ and $0<\gamma(s)\leq 1$~\cite{discount4}.\footnote{There are
alternatives to this tuning strategy.  \aij{For instance, \cite{bozkurt}
proposes state-dependent tuning where the reward function also depends on
the discount factor, while in this work the reward function is independent
of the discounting scheme.}}

\begin{definition}[Optimal Policy]\label{optimal_pol}
	An optimal policy $\pi^*$ is defined as follows:
	$$
	\pi^*(s) = \arg\sup\limits_{\pi \in \mathcal{D}}~ {U}^{\pi}(s),
	$$
	where $\mathcal{D}$ is the set of stationary \hos{deterministic} policies over the state space $\mathcal{S}$ defined below.
\end{definition}

\begin{definition}[Semi-Deterministic Policy]\label{def:semi_deterministic_policy}
\hos{We call a policy $\pi: \mathcal{S} \times \mathcal{A} \rightarrow
[0,1]$ semi-deterministic if for some states, say $s \in \bar{S} \subset \mathcal{S}$, the
policy $\pi$ is a uniform distribution over a set of actions $\varLambda(s)
\subseteq \mathcal{A}$, i.e.,  $\pi(s)=\mathit{unif}(\varLambda(s))$.  In
this work, whenever we talk about semi-deterministic policies, we assume
that at any state $s\in\mathcal{S}$ the set $\varLambda(s)$ is the set of
actions whose expected return is $U^{\pi^*}(s)$.  In many works this
specific interpretation of semi-determinism is simply referred to as policy
determinism.  However, a fully-deterministic policy is a special case of
semi-deterministic policies where $\varLambda(s)$ is a singleton for all
states $s \in \mathcal{S}$.}
\end{definition}

An MDP $\mathfrak{M}$ is said to be solved if the agent discovers an optimal
policy $\pi^*:\mathcal{S} \times \mathcal{A} \rightarrow [0,1]$ that
maximises the expected return.  We show later in the paper that synthesising
a policy whose traces satisfy an LTL specification with maximum probability
on MDP $\mathfrak{M}$ can be reduced to finding an optimal
\hos{semi-deterministic} policy that maximises an expected return on an
extended MDP~$\mathfrak{M}'$.  In the following we review the syntax and
semantics of LTL.

\subsection{Linear Temporal Logic Properties}\label{LTL}

LTL is a rich task specification language that allows to express a wide
range of properties (e.g., temporal, sequential, conditional).  In this
work, we use LTL specifications to formally and automatically shape a reward
function which, as we discuss later, would otherwise be tedious to express
and to achieve by conventional reward shaping methods.

\begin{definition}[Path] \label{def:path}
In an MDP $\mathfrak{M}$, an infinite path $\rho$ starting at $s_0$ is a
sequence of states $\rho= s_0 \xrightarrow{a_0} s_1 \xrightarrow{a_1} ... 
~$ such that every transition $s_i \xrightarrow{a_i} s_{i+1}$ is possible in
$\mathfrak{M}$, i.e.,  $s_{i+1}$ belongs to the smallest Borel set $B$ such
that $P(B|s_i,a_i)=1$ (or in a finite-state MDP, $s_{i+1}$ is such that
$P(s_{i+1}|s_{i},a_i)>0$).  We might also denote $\rho$ as $s_0..$ to
emphasise that $\rho$ starts from~$s_0$.
\end{definition}

Given a path $\rho$, the $i$-th state of $\rho$ is denoted by $\rho[i]$,
where $\rho[i]=s_{i}$.  Furthermore, the $i$-th suffix of $\rho$ is
$\rho[i..]$ where $\rho[i..]=s_i \xrightarrow{a_i} s_{i+1}
\xrightarrow{a_{i+1}} s_{i+2} \xrightarrow{a_{i+2}} s_{i+3}
\xrightarrow{a_{i+3}} ...~$. 
The language of LTL formulae over a given set of atomic propositions
$\mathcal{AP}$ is syntactically defined as~\cite{pnueli}
\begin{equation*}\label{ltlsyntax}
\varphi:= \mathit{true} ~|~ \alpha \in \mathcal{AP} ~|~ \varphi \land \varphi ~|~ \neg \varphi ~|~ \bigcirc \varphi ~|~ \varphi~ \mathrm{U}~ \varphi,
\end{equation*}
where the operators $ \bigcirc $ and $ \mathrm{U} $ are called ``next'' and
``until'', respectively.  The semantics of LTL formulae, as interpreted over
MDPs, is discussed in the following.

\begin{definition}[LTL Semantics]\label{def:ltl_semantics} 
For an LTL formula $\varphi$ and for a path $\rho$ in an MDP $\mathfrak{M}$,
the satisfaction relation $\rho\models\varphi$ is defined as~\cite{pnueli,bible}
\begin{equation*}
	\begin{aligned}
	& \rho \models \alpha \in \mathcal{AP} \Leftrightarrow \alpha \in L(\rho[0]), \\
	& \rho \models \varphi_1\wedge \varphi_2 \Leftrightarrow \rho \models \varphi_1\wedge \rho \models \varphi_2,\\
	& \rho \models \neg \varphi \Leftrightarrow \rho \not \models \varphi, \\
	& \rho \models \bigcirc \varphi \Leftrightarrow \rho[1..] \models \varphi, \\
	& \rho \models \varphi_1\mathrm{U} \varphi_2 \Leftrightarrow \exists j \in \mathbb{N}_0~\mbox{s.t.}~ \rho[j..] \models \varphi_2 ~\wedge 	
	~~\forall i,~0 \leq i < j,~ \rho[i..] \models \varphi_1.
	\end{aligned}
\end{equation*}
\end{definition}
\vspace{2mm}
The operator $\bigcirc$ (again, read as ``next'') requires  $\varphi$ to be
satisfied starting from the next-state suffix of the path $\rho$.  The
operator $\mathrm{U}$ (``until'') is satisfied over $\rho$ if
$\varphi_1$ continuously holds until $\varphi_2$ becomes $\mathit{true}$. 
By means of the until operator we are furthermore able to define two
temporal modalities:
(1) eventually, $\lozenge \varphi = \mathit{true} ~\mathrm{U}~ \varphi$; and
(2) always, $\square \varphi = \neg \lozenge \neg \varphi$.  The intuition
for $\lozenge \varphi$ is that $\varphi$ has to become $\mathit{true}$ at
some finite point in the future, whereas $\square \varphi$ means that
$\varphi$ has to remain $\mathit{true}$ forever.  An LTL formula $\varphi$
over $\mathcal{AP}$ specifies the following set of words:

\begin{equation}\label{words}
\mathit{Words}(\varphi)=\{\sigma \in (2^{\mathcal{AP}})^\omega ~\mbox{s.t.}~ \sigma \models \varphi\}.
\end{equation}

\rbtl{\begin{definition}[Safety Fragment of LTL]\label{def:safety}
Let us define an unsafe prefix of a set of words, e.g.,
$\mathit{Words}(\cdot)$, as a finite word $\sigma_{\mathit{finite}} \in
(2^{\mathcal{AP}})^*$ such that all infinite extensions, i.e.,
$\sigma_{\mathit{finite}}(2^{\mathcal{AP}})^\omega$, are not in
$\mathit{Words}(\cdot)$.  The safety fragment of LTL includes those formulae
whose violating words has an unsafe prefix.
\end{definition}}

\begin{definition}[Probability of Satisfying an LTL Formula] \label{ltlprobab} 
	Starting from any state $s$ and following a stationary semi-deterministic policy $\pi$, we denote the probability of satisfying formula~$\varphi$ as	
	$$
	\mathit{Pr}(s..^{\pi} \models \varphi),
	$$
	where $s..^{\pi}$ denotes the collection of all paths starting from state $ s $, generated under policy~$\pi$. The maximum probability of satisfaction is also defined as:
	$$
	\mathit{Pr}_{\max}(s_0 \models\varphi)=\sup\limits_{\pi \in \mathcal{D}} \mathit{Pr}(s_0..^{\pi} \models \varphi).
	$$
\end{definition}

Using an LTL formula we can now specify a set of constraints (i.e.,
requirements, or specifications) over the traces of the MDP.  Once a policy
$\pi$ is selected, it dictates which action has to be taken at each state of
the MDP~$\mathfrak{M}$, hence reducing the MDP to a Markov chain denoted
by~$\mathfrak{M}^{\pi}$.
For an LTL formula~$\varphi$, an alternative method to express the set
$\mathit{Words}(\varphi)$ in \eqref{words} is to employ a
Limit-Deterministic Generalised B\"uchi automaton (LDGBA)~\cite{sickert}. 
We first define a Generalised B\"uchi Automaton (GBA), then we formally
introduce the LDGBA~\cite{sickert}.
\begin{definition}[Generalised B\"uchi Automaton]\label{def:gba} 
A GBA $\mathfrak{A}=(\allowbreak\mathcal{Q},\allowbreak
q_0,\allowbreak\Sigma, \allowbreak\mathcal{F}, \allowbreak\Delta)$ is a
structure where $\mathcal{Q}$ is a finite set of states, $q_0 \in
\mathcal{Q}$ is the initial state, $\Sigma=2^{\mathcal{AP}}$ is a finite
alphabet, $\mathcal{F}=\{F_1,...,F_f\}$ is the set of accepting conditions,
where $F_j \subset \mathcal{Q}, 1\leq j\leq f$, and $\Delta: \mathcal{Q}
\times \Sigma \rightarrow 2^\mathcal{Q}$ is a transition
relation.
\end{definition}
Let $\Sigma^\omega$ be the set of all infinite words over $\Sigma$. An infinite word $\sigma \in \Sigma^\omega$ is accepted by a GBA $\mathfrak{A}$ if there exists an infinite run $\theta \in\mathcal{Q}^\omega$ starting from $q_0$ where $\theta[i+1] \in\Delta(\theta[i],\sigma[i]),~i \geq 0$ and for each $F_j \in \mathcal{F}$ 
\begin{equation} \label{acc}
\mathit{inf}(\theta) \cap F_j \neq \emptyset,
\end{equation}
where $\mathit{inf}(\theta)$ is the set of states that are visited infinitely often by the run $\theta$.

\begin{definition}[LDGBA]\label{ldbadef}
	A GBA $\mathfrak{A}=(\mathcal{Q},q_0,\Sigma, \mathcal{F}, \Delta)$ is limit-deterministic if $\mathcal{Q}$ can be partitioned into two disjoint sets $\mathcal{Q}=\mathcal{Q}_N \cup \mathcal{Q}_D$, such that~\cite{sickert}:
	\begin{itemize}
		\item \rbtl{$\mathcal{Q}_D$ is an invariant set:} $\Delta(q,\alpha) \subset \mathcal{Q}_D$ and $|\Delta(q,\alpha)|=1$ for every state $q\in\mathcal{Q}_D$ and for every $\alpha \in \Sigma$,
		\item for every $F_j \in \mathcal{F}$, $F_j \subset \mathcal{Q}_D$,

		\item $q_0 \in \mathcal{Q}_N$, and all the transitions from
                    $\mathcal{Q}_N$ to $\mathcal{Q}_D$ are non-deterministic
                    $\varepsilon$-transitions.
	\end{itemize}
  \rbtl{Unlike a standard transition in a GBA,
                    which requires a non-empty set of labels (see function $\Delta$ in
                    Definition~\ref{def:gba}), in an $\varepsilon$-transition this set can be
                    empty, which allows the automaton to change its state without reading any
                    atomic proposition.}
\end{definition}

Intuitively, an LDGBA is a GBA that has two partitions: initial
($\mathcal{Q}_N$) and accepting ($\mathcal{Q}_D$).  The accepting part
includes all the accepting states and has deterministic transitions. 
As an example, Fig.~\ref{fig:ldba_ex} shows the LDGBA constructed for
the formula $\varphi=a\wedge\bigcirc(\lozenge\square a\vee\lozenge\square
b)$.

\begin{figure}[!t]\centering
	\scalebox{1}{
		\begin{tikzpicture}[shorten >=1pt,node distance=2.7cm,on grid,auto] 
		\node[state,initial] (q_0)   {$q_0$}; 
		\node[state] (q_1) [right=of q_0]  {$q_1$}; 
		\node[state,accepting] (q_2) [above right=of q_1] {$q_2$}; 
		\node[state,accepting] (q_3) [below right=of q_1] {$q_3$};
		\node (N)[draw=blue, fit= (q_0)(q_1),dashed,ultra thick,inner sep=3mm] {};
		\node [yshift=2ex, blue] at (N.north) {$\mathcal{Q}_N$};
		\node (D)[draw=orange, fit= (q_2)(q_3),dashed,ultra thick,inner sep=3mm] {};
		\node [yshift=2ex, orange] at (D.north) {$\mathcal{Q}_D$}; 
		\path[->] 
		(q_0) edge node {$a$} (q_1)
		(q_1) edge  node {$\varepsilon$} (q_2)
		(q_1) edge [loop above] node {$\mathit{true}$} (q_1)
		(q_1) edge [below] node {$\varepsilon~~~$} (q_3)
		(q_2) edge  [loop right] node {$a$} (q_2)
		(q_3) edge  [loop right] node {$b$} (q_3);
		\end{tikzpicture}}
	\caption{LDGBA for the formula  $a\wedge\bigcirc(\lozenge\square a\vee\lozenge\square b)$.}
	\label{fig:ldba_ex}  
\end{figure}

\begin{remark}
The LTL-to-LDGBA algorithm used in this paper was proposed
in~\cite{sickert}.  It results in an automaton with two parts (initial
$\mathcal{Q}_N$ and accepting $\mathcal{Q}_D$), both of which use
deterministic transitions.  Additionally, there are non-deterministic
$\varepsilon$-transitions between them.  According to
Definition~\ref{ldbadef}, the discussed structure is still an LDGBA (the
determinism in the initial part is stronger than that required in the LDGBA
definition).  An $\varepsilon$-transition allows an automaton to change its
state without reading an input symbol.  \rbtl{In practice, during an episode
of LCRL algorithm, whenever the agent reaches the boundary of $
\mathcal{Q}_N $, e.g., state $q_1$ in Fig.~\ref{fig:ldba_ex}, the
$\varepsilon$-transitions between $ \mathcal{Q}_N $ and $ \mathcal{Q}_D $
reflect the agent's choices to move to $ \mathcal{Q}_D $.  The agent is free
to choose any of these transitions as they do not require the agent to read
any atomic proposition.  This is later clarified further in
Definition~\ref{def:product_mdp}.}
\end{remark}

\begin{definition}[Non-accepting Sink Component] \label{def:sinks}
\hos{A non-accepting sink component of the LDGBA
$\mathfrak{A}=(\mathcal{Q},q_0,\Sigma, \mathcal{F}, \Delta)$ is a directed
graph induced by a set of states $ Q \subset\mathcal{Q}$ such that (1) the
graph is strongly connected; (2) it does not include all accepting sets $
F_k,~k=1,...,f $; and (3) there exist no other strongly connected set $ Q'
\subset \mathcal{Q},~Q'\neq Q, $ such that $ Q \subset Q' $.  We denote the
union of all non-accepting sink components of $\mathfrak{A}$ as
$\mathds{N}$.}
\end{definition}

In the following we formally define the problem and discuss our proposed architecture Logically-Constrained Reinforcement Learning (LCRL). 

\section{Logically-Constrained Reinforcement Learning}\label{lcsec}

We are interested in synthesising a policy (or policies) for an unknown
(black-box) MDP via RL, such that the induced Markov chain satisfies a given
LTL property with maximum probability.  

\begin{assumption}
	In this paper, we assume that the MDP~$\mathfrak{M}$ is fully unknown, and the learning agent has no prior knowledge about the transition kernel $P$.~~\hfill$\lrcorner$
\end{assumption}

\aij{In the following, i}n order to explain the core
concepts of the algorithm and for ease of exposition, let us abstractly assume for now
that the MDP graph and the associated transition probabilities are known. 
Later these assumptions are entirely removed, and we stress that the
algorithm can be run model-free.  LCRL thus targets the issue of ``verified
learning'' at its core, namely the model-free learning-based synthesis of
policies that abide by a given LTL requirement.  Furthermore, in order to
handle the general case of non-ergodic MDPs, LCRL consists of several
resets, at each of which the agent is forced to re-start from the initial
state of the MDP: each reset defines an episode, as such the algorithm is
known as ``episodic RL''.

\begin{problem_1}
	\label{problem_definition}
	Given an MDP~$\mathfrak{M}$ and an LTL specification~$\varphi$, 
	we wish to find an optimal policy $\pi^*\in\mathcal{D}$ such that the probability of satisfying the specification from \green{any state} is maximised, i.e., $\pi^*~=~\arg\sup_{\pi \in \mathcal{D}} \mathit{Pr}({s}^{\pi}~\models~\varphi),\allowbreak~\forall s \in \mathcal{S}$.
	Furthermore, we would like to find the maximum probability of satisfaction $\mathit{Pr}_{\max}(s \models\varphi)$.~~\hfill$\lrcorner$
\end{problem_1}

In order to tackle Problem~\ref{problem_definition}, we first relate the MDP
model and the LDGBA constructed from the given LTL formula, by
``synchronising'' the two.  This new structure is, firstly, a good fit for
RL and, secondly, it generates a model that satisfies the given LTL
property.

\begin{definition} [Product MDP]\label{def:product_mdp}
	Given an MDP $\mathfrak{M}=(\allowbreak\mathcal{S},\allowbreak\mathcal{A},\allowbreak s_0,\allowbreak P,\allowbreak\mathcal{AP},L)$ and an LDGBA $\mathfrak{A}=(\mathcal{Q},q_0,\allowbreak\Sigma, \allowbreak\mathcal{F}, \allowbreak\Delta)$ with $\Sigma=2^{\mathcal{AP}}$, the product MDP is defined as $(\mathfrak{M}\otimes \mathfrak{A}) = \mathfrak{M}_\mathfrak{A}=(\mathcal{S}^\otimes,\allowbreak \mathcal{A}^\otimes,\allowbreak s^\otimes_0,P^\otimes,\allowbreak \mathcal{AP}^\otimes,\allowbreak L^\otimes,\allowbreak \mathcal{F}^\otimes)$, where $\mathcal{S}^\otimes = \mathcal{S}\times\mathcal{Q}$, \rbtl{$\mathcal{A}^\otimes = \mathcal{A}$}, $s^\otimes_0=(s_0,q_0)$, $\mathcal{AP}^\otimes = \mathcal{Q}$, $L^\otimes : \mathcal{S}^\otimes\rightarrow 2^\mathcal{Q}$ such that $L^\otimes(s,q)={q}$ and $\mathcal{F}^\otimes \subseteq {\mathcal{S}^\otimes}$ is the set of accepting states $\mathcal{F}^\otimes=\{F^\otimes_1,...,F^\otimes_f\}$, where ${F}^\otimes_j=\mathcal{S}\times F_j$. The transition kernel $P^\otimes$ is such that given the current state $(s_i,q_i)$ and action $a$, the new state is $(s_j,q_j)$, where $s_j\sim P(\cdot|s_i,a)$ and $q_j\in\Delta(q_i,L(s_j))$. When the MDP $\mathfrak{M}$ has a finite state space, then $P^\otimes:\mathcal{S}^\otimes \times \mathcal{A} \times \mathcal{S}^\otimes \rightarrow [0,1]$ is the transition probability function, such that $(s_i \xrightarrow{a} s_j) \wedge (q_i \xrightarrow{L(s_j)} q_j) \Rightarrow P^\otimes((s_i,q_i),a,(s_j,q_j))=P(s_i,a,s_j).$ Furthermore, in order to handle $\varepsilon$-transitions we make the following modifications to the above definition of product MDP:
	\begin{itemize}
		\item for every potential $\varepsilon$-transition to some state $q \in \mathcal{Q}$ we add a corresponding action $\varepsilon_q$ in the product:
		$$
		\mathcal{A}^\otimes=\mathcal{A}\cup \{\varepsilon_q, q \in \mathcal{Q}\}.
		$$
		
		\item the transition probabilities corresponding to $\varepsilon$-transitions are given by 
		\[\hspace{-5.5mm}P^\otimes((s_i,q_i),\varepsilon_q,(s_j,q_j)) = \left\{
		\begin{array}{lr}
		1 & $ if $  s_i=s_j,~q_i\xrightarrow{\varepsilon_q} q_j=q,\\
		0 & $ otherwise. $
		\end{array}
		\right.
		\]
	\end{itemize}
\end{definition}

Recall that an $\varepsilon$-transition between $ \mathcal{Q}_N $
and $ \mathcal{Q}_D $ corresponds in practical terms to a ``guess''
on reaching~$ \mathcal{Q}_D $.  \aij{The intuition behind the above
modification is that, during an exploration episode, once the
automaton state reaches the edge of $ \mathcal{Q}_N $ and an
$\varepsilon$-transition is necessary, the product automaton treats
this $\varepsilon$-transition as an extra action that can be taken
in the MDP.  Accordingly, during the RL exploration, if after an
$\varepsilon$-transition the associated labels in the accepting set
of the automaton cannot be read or the accepting states cannot be
visited, then the guess is deemed wrong, and the exploration in RL
is stopped.}

\begin{remark}
In order to clearly explain the role of different components in the
proposed approach, we have employed model-dependent notions, such as
transition probabilities and the product MDP.  However, we emphasise again
that the proposed approach can run ``model-free'', and as such that it does
not depend on these components.  In particular, as per
Definition~\ref{ldbadef}, the LDGBA is composed of two disjoint sets of
states $ \mathcal{Q}_D $ (which is invariant) and $ \mathcal{Q}_N $, where
the accepting states belong to the set $ \mathcal{Q}_D $.  Since all
transitions are deterministic within $ \mathcal{Q}_N $ and $ \mathcal{Q}_D
$, the automaton transitions can be executed simply by reading the labels,
which makes the agent aware of the automaton state without explicitly
constructing the product MDP.  We will later define a reward function
``on-the-fly'', emphasising that the agent does not need to know the model
structure or transition probabilities.
\end{remark}

Note that LTL is a temporal language and satisfying an LTL property requires
a policy that is possibly non-Markovian and has an embedded
memory~\cite{camacho,camacho2}.  By constructing the product MDP we add an
extra dimension to the state space of the original MDP, namely the states of
the automaton representing the LTL formula.  The role of the added dimension
is to track LTL satisfaction and, hence, to synchronise the current state of
the MDP with the state of the automaton: this essentially converts the
non-Markovian LTL policy synthesis problem over the original MDP to a
Markovian one over the product MDP.

In the following we briefly explain how a non-Markovian task can be broken
down into simple composable Markovian sub-tasks (or modules).  Each state of
the automaton in the product MDP (Definition~\ref{def:product_mdp}) is a
``task divider'' and each transition between these states is a ``sub-task''. 
For example consider a sequential task of visit $a$ and then $b$ and finally
$c$, i.e.,
$$
\Diamond(a\wedge\Diamond (b \wedge\Diamond c)).
$$
The corresponding automaton for this LTL task is given in Fig.~\ref{exam}. 
The entire task is modularised into three sub-tasks, i.e., reaching $a$,
$b$, and then $c$, and each automaton state acts as a divider.  For each
automaton state $q_i$, RL needs to focus only on the outgoing edges of
$q_i$.  For instance, at $q_2$ in Fig.~\ref{exam}, RL only needs to find a
policy whose traces satisfy the sub-formula~$\Diamond b$. 
Fig.~\ref{fig:product_mdp_ex} illustrates the construction of a product MDP
with the generated LDGBA in Fig.~\ref{fig:ldba_ex}.

\begin{figure}[!t]\centering
	\scalebox{1}
	{
		\begin{tikzpicture}[shorten >=1pt,node distance=2cm,on grid,auto] 
			\node[state,initial] (q_1)   {$q_1$}; 
			\node[state] (q_2) [right=of q_1] {$q_2$};
			\node[state] (q_3) [right=of q_2] {$q_3$};
			\node[state,accepting] (q_4) [right=of q_3] {$q_4$};
			\path[->] 
			(q_1) edge [loop above] node {$\neg a$} ()
			(q_1) edge  node {$a$} (q_2)
			(q_2) edge [loop above] node {$\neg b$} ()
			(q_2) edge node {$b$} (q_3)
			(q_3) edge [loop above] node {$\neg c$} ()
			(q_3) edge node {$c$} (q_4)
			(q_4) edge [loop above] node {$c$} ();
		\end{tikzpicture}
	}
	\caption{LDGBA for a sequential mission task.}
	\label{exam}
\end{figure}

\setlength{\fboxrule}{0pt}
\begin{figure}[!t]\centering
	\subfloat[]{{
			\scalebox{1}{
				\begin{tikzpicture}[shorten >=1pt,node distance=1.8cm,on grid,auto] 
					\node[state,initial] (q_0)   {$q_0$}; 
					\node[state] (q_1) [right=of q_0]  {$q_1$}; 
					\node[state,accepting] (q_2) [above right=of q_1] {$q_2$}; 
					\node[state,accepting] (q_3) [below right=of q_1] {$q_3$};
					\path[->] 
					(q_0) edge node {$a$} (q_1)
					(q_1) edge  node {$\varepsilon$} (q_2)
					(q_1) edge [loop above] node {$\mathit{true}$} (q_1)
					(q_1) edge [below] node {$\varepsilon~~~$} (q_3)
					(q_2) edge  [loop right] node {$a$} (q_2)
					(q_3) edge  [loop right] node {$b$} (q_3);
				\end{tikzpicture}
			}
	}}
	\subfloat[]{{
			\scalebox{1}{
				\begin{tikzpicture}[shorten >=1pt,node distance=2cm,on grid,auto] \node[state,initial,label=below:$\{a\}$] (s_0) {$s_0$};
					\node[state,label=below:$\{b\}$] (s_1) [right=of s_0]{$s_1$};
					\draw[->] (s_0) -- node [below] {$0.9$} (s_1);
					\draw[->] (s_0) [out=30,in=80,loop] to coordinate[pos=0.2](aa) node [above] {$\fbox{0.1}$} (s_0);
					\draw[->] (s_1) [out=30,in=80,loop] to coordinate[pos=0.2] node [above] {\fbox{$\textcolor{red}{a_2}:1$}} (s_1);
					\path pic[draw, angle radius=8mm,"$\textcolor{red}{a_1}$",angle eccentricity=1.4] {angle = s_1--s_0--aa};
				\end{tikzpicture}
			}
	}}
	$\qquad$
	\subfloat[]{{
			\scalebox{0.75}{
				\begin{tikzpicture}[shorten >=1pt,node distance=3cm,on grid,auto] 
					\node[state,initial] (q_0)   {$(s_0,q_0)$}; 
					\node[state] (q_1) [right=of q_0]  {$(s_1,q_1)$}; 
					\node[state] (q_2) [right=of q_1] {$(s_1,q_2)$}; 
					\node[state,accepting] (q_3) [below right=of q_1] {$(s_1,q_3)$};
					\draw[->] (q_0) -- node [below] {$0.9$} (q_1);
					\draw[->] (q_0) [out=30,in=80,loop] to coordinate[pos=0.2](aa) node [above] {$\fbox{0.1}$} (q_0);
					\draw[->] (q_1) [out=30,in=80,loop] to coordinate[pos=0.2] node [above] {\fbox{$\textcolor{red}{a_2}:1$}} (q_1);
					\draw[->] (q_1) -- node [below] {$\textcolor{red}{\varepsilon_{q_2}}$} (q_2);
					\draw[->] (q_1) -- node [below] {$\textcolor{red}{\varepsilon_{q_3}}~~$} (q_3);
					\path pic[draw, angle radius=14mm,"$\textcolor{red}{a_1}$",angle eccentricity=1.3] {angle = q_1--q_0--aa};
					\draw[->] (q_3) [out=335,in=25,loop] to coordinate[pos=0.2] node [above] {$~~~~~~~~~~~\textcolor{red}{a_2}:1$} (q_3);
				\end{tikzpicture}
			}
	}}
	\caption{Example of product MDP: (a) the LDGBA from Fig.~\ref{fig:ldba_ex} and (b) an MDP; (c) the product of the MDP and the LDGBA, as per Definition~\ref{def:product_mdp}. \hos{We rely on the observation \cite{sickert2} that it is sufficient to take $\varepsilon$-transitions only from states in the max-end components of the product of $\mathfrak{M}$ and the initial partition of the LDGBA $\mathcal{Q}_N$. Hence no $\varepsilon$-transitions have to be produced in the initial state of $\mathfrak{M}_\mathfrak{A}$}.}
	\label{fig:product_mdp_ex}
\end{figure}

\aij{The product MDP (e.g., Fig.~\ref{fig:product_mdp_ex}) provides a
structure that allows us to shape a reward function for an RL algorithm, by
leveraging the accepting condition of the LDGBA.  Such a reward function
thus clearly relates to the satisfaction of the given LTL formula.} Before
introducing such a reward assignment, we need to define the ensuing
function.  Recall that a generalised B\"uchi automaton accepts words that
visit its accepting sets infinitely often.  \rbtl{The role of the following
function is to track and output the subset of accepting sets that need to be
visited at any given time.  Namely, during the learning process, we would
like to know precisely the set of labels that ought to be read (possibly
once more), so that by such repeated visitations the specified LTL task is
eventually satisfied.}

\begin{definition}[Accepting Frontier Function] \label{frontier}
\rbtl{Let $\mathfrak{A}=(\allowbreak\mathcal{Q},\allowbreak q_0,\allowbreak\Sigma,\allowbreak\mathcal{F},\allowbreak\Delta)$, be an LDGBA where $\mathcal{F}=\{F_1,...,F_f\}$ is the set of accepting conditions, and $F_j \subset \mathcal{Q}, 1\leq j\leq f$. Define the function $ Acc:\mathcal{Q}\times \mathcal{F}\rightarrow2^\mathcal{Q} $ as the accepting frontier function, which executes the following operation over a given set $ \mathds{F}\subset \mathcal{F}$ for every $F_j\in\mathcal{F}$: 	
	\[Acc(q,\mathds{F})=\left\{
	\begin{array}{lr}
	\mathds{F}{\setminus F_j}~~~ & (q \in F_j) \wedge (\mathds{F}\neq \{F_j\}),\\
	\\
	{\mathcal{F}}{\setminus F_j} ~~~ & (q \in F_j) \wedge (\mathds{F}=\{F_j\}) \wedge (\mathcal{F}{\setminus F_j} \neq \emptyset),\\
	\mathds{F} & $otherwise.$
	\end{array}
	\right.
	\] 
	Once state $ q\in F_j $ and set $ \mathds{F} $ are fed to function $ Acc $, it outputs a set containing the elements of $ \mathds{F} $ minus $ F_j $. 
	However, if $ \mathds{F}=F_j $, then the output is the family of 
	all accepting sets of the LDGBA minus the set $ F_j $.  Finally, if
        the state $ q $ is not an accepting state, then the output of $ Acc $ is $
        \mathds{F} $.  In short, the accepting frontier function excludes from
        $\mathds{F}$ the accepting set that is currently visited, unless it is the
        only remaining accepting set.  Otherwise, the output of $Acc(q,\mathds{F})$
        is $\mathds{F}$ itself.  What remains in $\mathds{F}$ are those accepting
        sets that still need to be visited in order to attain the generalised
        B\"uchi accepting condition, as per
        Definition~\ref{def:gba}.}
\end{definition}

\aij{As discussed before, t}he product MDP encompasses the transition relations of the original MDP and
the structure of the B\"uchi automaton, and it inherits characteristics of
both.  Thus, a proper reward function leads the RL agent to find a policy
that is optimal, in the sense that it satisfies the LTL property~$\varphi$
with maximal probability.  We introduce an on-the-fly reward function that fits
the model-free RL architecture: when an agent observes the current state $
s^\otimes $, implements action $ a $ and observes the subsequent state $
{s^\otimes}' $, the agent is given a scalar reward, as follows:
\begin{equation}\label{thereward}
\begin{aligned}
R(s^\otimes,a) = \left\{
\begin{array}{lr}
r_p & $ if $  q' \in \mathds{A},~{s^\otimes}'=(s',q'),\\
r_n & $ otherwise.$
\end{array}
\right.
\end{aligned}
\end{equation} 
Here, $r_p > 0$ is a positive reward and $r_n = 0$ is a neutral reward.  A
positive reward is assigned to the agent when it takes an action that leads
to a state with a label in $\mathds{A}$.  The set $ \mathds{A} $ is called
the accepting frontier set, is initialised as the family of sets \rbtl{$
\mathds{A}=\{F_k\}_{k=1}^{f} =\mathcal{F}$}, and is updated by the following
rule every time after the reward function is evaluated:
$$
\mathds{A}\leftarrow Acc(q',\mathds{A}). 
$$
\rbtl{The set $ \mathds{A} $ always contains the set of accepting states
that ought to be visited at any given time: in this sense the reward
function is ``synchronous'' with the accepting condition set by the LDGBA. 
Thus, the agent is guided by the above reward assignment to visit those
states and once all the sets $ F_k,~k=1,...,f, $ are visited, the accepting
frontier $ \mathds{A} $ is reset.  As a consequence, as the agent
continuously explores, it is guided to visit all the accepting sets
infinitely often and is rewarded towards the satisfaction of the
corresponding LTL property.  Considering the syntactic tree of the LTL
property, by visiting each accepting set $F_k$ infinitely often, the agent
satisfies the corresponding sub-formula of the LTL formula.  This means
that, by guiding the agent to visit all the accepting sets, we are
essentially guiding the agent to move upwards in the syntactic tree towards
the satisfaction of the entire LTL property.  We elaborate on the issue of
partial satisfaction of the LTL property in the supplementary material
(Appendix~\ref{appndx:partial_sat}).}

The reward structure depends on parameters $r_p=M+y\times m\times
\mathit{rand}(s^\otimes)$ and $r_n=y \times m \times
\mathit{rand}(s^\otimes)$.  The parameter $y\in\{0,1\}$ is a constant, $0 <
m \ll M$ are arbitrary positive values, and $\mathit{rand}:
\mathcal{S}^\otimes \rightarrow (0,1)$ is a function that generates a random
number in $(0,1)$ for each state $s^\otimes$ each time $ R $ is being
evaluated.  The role of the function $rand$ is to \aij{resolve possible symmetry issues\footnote{If all weights in a feedforward net start with
		equal values and if the solution requires unequal weights be learnt, the
		neural network might not proceed to learn.  Identical weights within the same
		hidden layer induce symmetries, which the neural net must break in order to
		generalise, reduce redundancies on the weights, and optimise the loss
		function~\cite{apprx}.}} when neural
nets are used for approximating the Q-function (namely, when the MDP state
space is continuous).  Also, note that parameter $y$ acts as a switch to
bypass the effect of the $rand$ function on $ R $ when no neural net is
used.  Thus, this switch is active $y=1$ when the MDP state space is
continuous, and disabled in other cases $y=0$.

\begin{remark}\label{on_the_fly_remark}
As our implementation is model-free, note that when running the proposed
algorithm there is no need to ``explicitly build'' the product MDP and to
store all its states and transitions in memory.  The automaton transitions
can be executed on-the-fly as the agent reads the labels of the MDP states. 
Namely, the agent can track the automaton state by just looking at the trace
that has been read so far.  The agent only needs to store the current state
of the automaton and observe the label at each step to check whether the
automaton state has changed or not.
\end{remark} 

\aij{In the following, we further elaborate on how the automaton state tracks the evolution of the accepting frontier set $\mathds{A}$.}

\begin{proposition}\label{g_monitor_proposition}
Given an LTL formula $\varphi$ and its associated LDGBA
$\mathfrak{A}=\allowbreak(\mathcal{Q},q_0,\allowbreak\Sigma,\allowbreak
\mathcal{F},\allowbreak \Delta)$, the set members of  $ \mathds{A} $ only
depend on the current state of the automaton and not on the sequence of
automaton states that have been already visited.  (Proof in the supplementary material.)
\end{proposition}

\hos{Proposition~\ref{g_monitor_proposition}
allows us to reason about the evolution of the accepting frontier set $\mathds{A}$ throughout the learning process. Specifically, 
the accepting sets are removed from $\mathds{A}$ upon visiting accepting
sets in $\mathcal{Q}_D$ until all accepting sets are visited at least once. 
This resets the accepting frontier set $\mathds{A}$
(Definition~\ref{frontier}), and thus the agent receives a positive reward
infinitely often for visiting accepting sets as specified in
\eqref{thereward}}.  \aij{Given this reward structure, we show in the
following that the optimal policy generated by an RL scheme maximises, in
the limit, the probability of satisfying the LTL property.}

\begin{thm}\label{thm:1}
Let $\varphi$ be the given LTL property and $\mathfrak{M}_\mathfrak{A}$ be
the product MDP constructed by synchronising the MDP~$\mathfrak{M}$ and the
LDGBA~$\mathfrak{A}$ expressing $\varphi$.  There exists a discount factor
that is close enough to $1$ under which an optimal Markov policy on
$\mathfrak{M}_\mathfrak{A}$ that maximises the expected return over the
reward in \eqref{thereward} also maximises the probability of satisfying
$\varphi$.  This optimal Markov policy induces a finite-memory policy on the
MDP~$\mathfrak{M}$.  (Proof in the supplementary material.)
\end{thm}

\begin{remark}
Note that the projection of policy $\pi^*$ onto the state space of
the original MDP~$\mathfrak{M}$ yields a finite memory policy
$\pi^*_\mathfrak{M}$.  Thus, if the generated traces under $\pi^*$ maximise
the LTL satisfaction probability, so do the traces under
$\pi^*_\mathfrak{M}$.
\end{remark}

\begin{remark}
\hos{The optimality of the policies generated using (deep)
neural-network-based approaches depends on a number of factors, such as the
network structure, number of hidden layers, and activation functions. 
Specifically, convergence guarantees for such methods to a true optimal
policy are not well developed and quantification of the sub-optimality of the
policy generated by these methods is out of the scope of this work.}
\end{remark}

An interesting extension of Theorem \ref{thm:1} is the following corollary.
\begin{corollary}[Maximum Probability of Satisfaction]\label{cor:max_sat_probab}
	From Definition \ref{ltlprobab}, for a discounting factor close enough to $1$ the maximum probability of satisfaction at any state $s^\otimes$ can be determined from the LCRL value function as 
	$$
	\mathit{Pr}_{\max}(s^\otimes \models\varphi) = \dfrac{1-\eta}{r_p}~ {U}^{{\pi}^*}(s^\otimes). 
	$$ (Proof in the supplementary material.)
\end{corollary}

\begin{remark}
	\hos{As mentioned in Definition \ref{def:semi_deterministic_policy}, semi-deterministic and deterministic policies are used interchangeably in a number of works. However, in general, at any state $s^\otimes$, there might be multiple actions $\varLambda(s^\otimes)$ whose expected return is ${U}^{{\pi}^*}(s^\otimes)$. An optimal action is then uniformly selected from $\varLambda(s^\otimes)$ as per Definition~\ref{def:semi_deterministic_policy}. Let us assume for the sake of simplicity that the action space is finite and we expand the definition of the expected utility from Definition~\ref{def:expected_utility}:
	\begin{align}\label{semi_det_sh}
		\begin{aligned}
		\mathit{Pr}_{\max}(s^\otimes \models\varphi)& = \dfrac{1-\eta}{r_p} \times\\
		&\sum_{a'\in\varLambda(s^\otimes)}\dfrac{1}{|\varLambda(s^\otimes)|}\mathds{E}^{\pi^*} \Big[\sum\limits_{n=0}^{\infty} \gamma(s^\otimes_n)^{N(s^\otimes_n)}~ r_n \Big| s^\otimes_0=s^\otimes,a_0=a'\Big],
		\end{aligned}
	\end{align}
	where $1/|\varLambda(s^\otimes)|$ is the probability of either of actions in $\varLambda(s^\otimes)$ to be selected. Since all the actions in $\varLambda(s^\otimes)$ have the same expected return ${U}^{{\pi}^*}(s^\otimes)$, the conditional expectation on the RHS of \eqref{semi_det_sh} is independent of $a'$. Thus, it can be factored out from the summation:
	\begin{equation*}\label{semi_det_sh_2}
	\mathit{Pr}_{\max}(s^\otimes \models\varphi) = \dfrac{1-\eta}{r_p}~{U}^{{\pi}^*}(s^\otimes)\sum_{a'\in\varLambda(s^\otimes)}\dfrac{1}{|\varLambda(s^\otimes)|}.
	\end{equation*}
	Note that $\sum_{a'\in\varLambda(s^\otimes)}{1}/{|\varLambda(s^\otimes)|}=1$, which means that $\pi^*$ maximises the satisfaction probability even when the optimal policy is semi-deterministic. In case when the action space is not finite, the summation in \eqref{semi_det_sh} is an integral.}
\end{remark}

\begin{thm}\label{thm:complexity}
\rbtl{Let $n$ be the length of the formula $\varphi$, i.e., the size of its
syntactic tree or the number of $\bigcirc$ and $\mathrm{U}$ operators used
in $\varphi$ as per Definition~\ref{def:ltl_semantics}.  Then for any LTL
formula in the LTL$\setminus$GU fragment, there exist an LDGBA with size of
only $2^{\mathcal{O}(n)}$ while an equivalent DRA has a size of
$2^{2^{\mathcal{O}(n)}}$ \cite{sickert2,kini2}.}
\end{thm}

\rbtl{It is easy to find examples similar to the LTL$\setminus$GU
fragment~\cite{sickert} in addition to efficient procedures for LDGBA
complementation~\cite{blahoudek2016complementing}.  As per
Theorem~\ref{thm:complexity}, LDGBAs are much more succinct than DRAs, and
hence in LCRL, the space over which the learning is performed (i.e.,
$\mathfrak{M}_\mathfrak{A}$) is $\mathcal{O}(|\mathcal{S}||\mathcal{A}|)
2^{\mathcal{O}(n)}$.  Given that in principle an RL agent has to visit all
the state-action pairs of the learning space to converge~\cite{sutton}, this
exponential reduction in the size of $\mathfrak{M}_\mathfrak{A}$
significantly improves convergence speed and sample efficiency.  Note that,
in the worst case, LTL to LDGBA translation is doubly-exponential,
which is just as expensive as LTL to DRA, and therefore, LDBGAs are
in practice smaller than DRAs.}

As a concluding remark, notice that LCRL is a general policy synthesis
architecture that is adaptable to many off-the-shelf model-free RL
algorithms.  The LCRL \hos{RL Engine} in Fig.~\ref{fig:overview} is a good
fit for a broad variety of RL schemes that conform with the required state
and action space cardinality and dimension.  Within the LCRL architecture,
the MDP and LDGBA states are synchronised, resulting in an on-the-fly
product MDP.  In the following sections we then discuss the applicability of
LCRL, and we present case studies to demonstrate the ease of use, and
scalability of the scheme.

\section{Logically-Constrained Tabular RL}

Recall that the simplest case in Definition~\ref{def:general_mdp} is when
the MDP state space and action space are both finite.  This case,
however, covers a significant number of control applications.  In this
section we discuss and show LCRL, a model-free RL architecture with
discounted reward, can be efficiently employed for LTL policy synthesis and
quantitative model checking in finite MDPs.

Q-learning (QL) is the most extensively used RL algorithm for synthesising
optimal policies in finite-state MDPs~\cite{sutton}.  In this section, the
LCRL \hos{RL Engine} in Fig.~\ref{fig:overview} is QL: we run QL over the
product MDP $\mathfrak{M}_\mathfrak{A}$ with the reward shaping proposed in
\eqref{thereward}, where we have set $ y=0 $.  In order to also handle
non-ergodic MDPs, we propose to employ a variant of standard QL that
consists of several resets, at each of which the agent is forced to re-start
from its initial state $ s_0 $.  Each reset defines an episode, and hence the
algorithm is called ``episodic QL''.  However, for the sake of brevity,
we omit the term ``episodic'' in the rest of the paper and we use the
term Logically-Constrained QL (LCQL).

For each state $s^\otimes \in \mathcal{S}^\otimes$ and for any available
action $a \in \mathcal{A}^\otimes$, QL assigns a quantitative value
$Q:\mathcal{S}^\otimes\times\mathcal{A}^\otimes\rightarrow \mathbb{R}$,
which is initialised with an arbitrary and finite value over all
state-action pairs.  As the agent starts receiving rewards and learning, the
Q-function is updated by the following rule for taking action $ a $ at state
$ s^\otimes $:

\begin{align}\label{ql_update_rule}
\begin{aligned}
Q(s^\otimes,a) \leftarrow& (1-\mu)Q(s^\otimes,a)+ 
\mu\big[R(s^\otimes,a)+\gamma(s^\otimes) \max\limits_{a' \in \mathcal{A}^\otimes}(Q({s'}^\otimes,a'))],
\end{aligned}
\end{align}
where $ Q(s^\otimes,a) $ is the Q-value corresponding to state-action $
(s^\otimes,a) $, $ 0<\mu\leq 1 $ is called learning rate or step size, $
R(s^\otimes,a) $ is the reward obtained for performing action $a$ in state
$s$, $\gamma$ is the discount factor, and ${s'}^\otimes$ is the state
obtained after performing action $a$.  The Q-function for the rest of the
state-action pairs remains unchanged.

Under mild assumptions over the learning rate~\cite{NDP,watkins}, for
finite-state and -action spaces QL converges to a unique limit.  This unique
limit is the expected discounted reward by taking action $ a $ at state $ s
$, and following the optimal policy afterwards.  Let us call this limit
$Q^*$.  Once QL converges, an optimal policy $\pi^*:
\mathcal{S}^\otimes \rightarrow \mathcal{A}^\otimes$ can be generated by
selecting the action that yields the highest $Q^*$, i.e.,
$$
\pi^*(s^\otimes)\in\arg\max\limits_{a \in \mathcal{A}}~Q^*(s^\otimes,a). 
$$
Here $ \pi^* $ corresponds to the optimal policy that can be generated via
DP.  This means that when QL converges, we have

\begin{equation*}\label{qnv}
Q^*(s^\otimes,a)=R(s^\otimes,a)+\gamma(s^\otimes)\sum\limits_{{s'}^\otimes\in\mathcal{S}^\otimes} P(s^\otimes,a,{s'}^\otimes) U^{\pi^*}({s'}^\otimes), 
\end{equation*}
where $ {s'}^\otimes $ is the agent new state after choosing action $a$ at state $s^\otimes$ such that $P({s'}^\otimes|s^\otimes, a)>0$.

\begin{algorithm2e}[!t]
	\DontPrintSemicolon
	\SetKw{return}{return}
	\SetKwRepeat{Do}{do}{while}
	\SetKwFunction{terminate}{episode$\_$terminate}
	\SetKwFor{terminatedef}{episode$\_$terminate()}{}{}
	\SetKwData{conflict}{conflict}
	\SetKwData{safe}{safe}
	\SetKwData{sat}{sat}
	\SetKwData{unsafe}{unsafe}
	\SetKwData{unknown}{unknown}
	\SetKwData{true}{true}
	\SetKwData{false}{false}
	\SetKwInOut{Input}{input}
	\SetKwInOut{Output}{output}
	\SetKwFor{Loop}{Loop}{}{}
	\SetKw{KwNot}{not}
		\Input{LTL specification, $ \textit{it\_threshold} $, $ \gamma $, $ \mu $}
		\Output{$\pi^*$}
		initialize $Q: \mathcal{S}^\otimes \times \mathcal{A}^\otimes \rightarrow \mathbb{R}^+_0$\;
		convert the desired LTL property to LDGBA $\mathfrak{A}$\;
		initialize $ \mathds{A} $\;
		initialize $episode$-$number:=0$\;
		initialize $iteration$-$number:=0$\;
		\While{$Q$ is not converged}
		{
			$episode$-$number++$\;
			$s^\otimes=(s_0,q_0)$\;
			\While{$ (q \notin {\mathds{N}}:~s^\otimes=(s,q))~ \& ~ (iteration$-$number<\text{it\_threshold})$}
			{
				$iteration$-$number++$\;
				choose  $a_*=\pi({s^\otimes})\in\arg\max_{a \in \mathcal{A}} Q({s^\otimes},a)$ $ ~\#$~$ \epsilon-$greedy or softmax are applicable\;
				move to $s^\otimes_*=(s_*,q_*)$ by $a_*$\; 
				receive the reward $R({s^\otimes},a_*)$\;
				$\mathds{A}\leftarrow Acc(q_*,\mathds{A})$\;
				$Q({s^\otimes},a_*)\leftarrow Q({s^\otimes},a_*)+\mu[R({s^\otimes},a_*)-Q({s^\otimes},a_*)+\gamma(s^\otimes) \max_{a'}(Q(s^\otimes_*,a'))]$\;
				$s^\otimes=s^\otimes_*$\;
			}
		}
	\caption{Logically-Constrained QL}
	\label{alg:lcql}
\end{algorithm2e} 

\section{Logically-Constrained Neural Fitted Q-iteration}

LCQL is focused on problems in which the set of states of
the MDP and the set of possible actions are both finite. Nonetheless, many
interesting real-world problems require actions to be taken in response to
high-dimensional or real-valued states~\cite{doya}. 
We can collect a number of samples and only then apply an approximation
function that is constructed via regression over the set of samples.  The
approximation function essentially replaces the conventional LCQL
state-action-reward look-up table by generalising over the state space of
the MDP.  In this section, we extend the LCRL architecture to a model-free RL
algorithm based on Neural Fitted Q-iteration (NFQ), which can synthesise an
optimal policy for an LTL property when the given MDP has a continuous
state space.  \hos{We replace the QL algorithm in LCRL \hos{RL
Engine} in Fig.~\ref{fig:overview} by NFQ to showcase the flexibility of
the LCRL architecture.} We call this algorithm Logically-Constrained NFQ
(LCNFQ) and we show that the proposed architecture is efficient and is
compatible with RL algorithms that are core of recent developments in the
community.  We have studied a number of alternative RL-based approaches to
LCNFQ in \cite{lcnfq}, and LCNFQ easily outperformed the competitors.

NFQ is an algorithm that employs feedforward neural
networks~\cite{multilayer} to approximate the Q-function, namely to
efficiently generalise or interpolate it over the entire state space,
exploiting a finite set of experience samples.  This set is called
experience replay.  Instead of the conventional QL update rule
in~\eqref{ql_update_rule}, NFQ introduces a loss function that measures the
error between the current Q-values $Q(s,a)$ and their target value
$R(s,a)+\gamma \max\limits_{a'}Q(s',a')$, namely
\begin{equation}\label{loss_function}
L=[Q(s,a)-R(s,a)+\gamma \max\limits_{a'}Q(s',a')]^2.
\end{equation}

In LCNFQ, the experience replay method is adapted to the product MDP
structure, over which we let the agent explore the MDP and reinitialise it
when a positive reward is received or when no positive reward is received
after a given number $ \mathit{th} $ of iterations.  The parameter $
\mathit{th} $ is set manually according to the state space of the MDP,
allowing the agent to explore the MDP while keeping the size of the sample
set limited.  All the traces that are gathered within episodes, i.e., 
experiences, are stored in the form of
$(s^\otimes,a,{s^\otimes}',R(s^\otimes,a),q)$, where $s^\otimes=(s,q)$ is
the current state in the product MDP, $a$ is the selected action,
${s^\otimes}'=(s',q')$ is the subsequent state, and $R(s^\otimes,a)$ is the
reward gained as in \eqref{thereward} with $y=1$ in~$r_p$.  The set of past
experiences is called the sample set $\mathcal{E}$.

Once the exploration phase is completed and the sample set is created,
learning is performed over the sample set.  In the learning phase, we
propose a hybrid architecture of $n$ separate feedforward neural nets, each
with one hidden layer, where $n=|\mathcal{Q}|$ and $\mathcal{Q}$ is the
finite cardinality of the automaton~$ \mathfrak{A} $\footnote{Different
embeddings, such as the one-hot encoding~\cite{onehot} and the integer
encoding, have been applied in order to approximate the global Q-function
with a single feedforward net.  \rbtl{However, we have observed poor performance
since these encodings allow the network to assume relationships
between automaton states that might not agree with the automaton structure.}  Clearly, this disrupts Q-function generalisation by assuming
that some states in product MDP are closer to each other.  Consequently, we
have turned to the use of $n$ separate neural nets, which work together in a
hybrid fashion, meaning that the agent can switch between these neural nets
as it jumps from one automaton state to another.}.  Each neural net is
associated with a state in the LDGBA and for each automaton state $q_i \in
\mathcal{Q}$ the associated neural net is called
$B_{q_i}:\mathcal{S}^\otimes\times\mathcal{A}\rightarrow\mathbb{R}$.  Once
the agent is at state $s^\otimes=(s,q_i)$, the neural net $B_{q_i}$ is used
for the local Q-function approximation.  The set of neural nets acts as a
global hybrid Q-function approximator $
Q:\mathcal{S}^\otimes\times\mathcal{A}\rightarrow\mathbb{R} $.  Note that
the neural nets are not fully decoupled.  For example, assume that by taking
action $a$ in state $s^\otimes=(s,q_i)$ the agent is moved to state
${s^\otimes}'=(s',q_j)$ where $ q_i\neq q_j $.  According to
(\ref{loss_function}) the weights of $B_{q_i}$ are updated such that
$B_{q_i}(s^\otimes,a)$ has minimum possible error to
$R(s^\otimes,a)+\gamma\max_{a'} B_{q_j}({s^\otimes}',a')$.  Therefore, the
value of $B_{q_j}({s^\otimes}',a')$ affects $B_{q_i}(s^\otimes,a)$.

\begin{algorithm2e}[!t]
	\DontPrintSemicolon
	\SetKw{return}{return}
	\SetKwRepeat{Do}{do}{while}
	\SetKwData{conflict}{conflict}
	\SetKwData{safe}{safe}
	\SetKwData{sat}{sat}
	\SetKwData{unsafe}{unsafe}
	\SetKwData{unknown}{unknown}
	\SetKwData{true}{true}
	\SetKwInOut{Input}{input}
	\SetKwInOut{Output}{output}
	\SetKwFor{Loop}{Loop}{}{}
	\SetKw{KwNot}{not}
		\Input{the set of experience samples $\mathcal{E}$}
		\Output{approximated Q-function}
		initialise all neural nets $B_{q_i}$ with $ (s_0,q_i,a) $ as the input and $ r_n $ as the output where $ a\in\mathcal{A} $ is a random action\;
		\Repeat{end of trial}
		{
			\For{$q_i=|\mathcal{Q}|$ \textbf{to} $1$}
			{
				$\mathcal{P}_{q_i}=\{(input_l,target_l),~l=1,...,|\mathcal{E}_{q_i}|)\}$\;
				~~~~~~~~~$input_l=({s_l}^\otimes,a_l)$\;
				~~~~~~~~~$target_l=R({s_l}^\otimes,a_l)+\gamma \max \limits_{a'} Q({{s_l}^\otimes}',a')$\;
				~~~~~~~~~where $({s_l}^\otimes,a_l,{{s_l}^\otimes}',R({s_l}^\otimes,a_l),{q_i}) \in \mathcal{E}_{q_i}$\;
				$B_{q_i} \leftarrow$ Rprop$(\mathcal{P}_{q_i})$
			}
		}
	\caption{Logically-Constrained NFQ}
	\label{lcnfqal}
\end{algorithm2e}

Let $q_i \in \mathcal{Q}$ be a state in the LDGBA.  Then define
$\mathcal{E}_{q_i}:=\{(\cdot,\cdot,\cdot,\cdot,x) \allowbreak \in
\mathcal{E} | x = q_i \}$ as the set of experiences within $\mathcal{E}$
that are associated to state $q_i$, i.e., $\mathcal{E}_{q_i}$ is the
projection of $\mathcal{E}$ onto $q_i$.  Once the exploration phase is
completed, each neural net $B_{q_i}$ is trained based on the associated
experience set $\mathcal{E}_{q_i}$.  At each iteration of training, a
pattern set $\mathcal{P}_{q_i}$ is generated based on the experience set
$\mathcal{E}_{q_i}$:
$$
\mathcal{P}_{q_i}=\{(input_l,target_l), l=1,...,|\mathcal{E}_{q_i}|)\},
$$
where $input_l=({s_l}^\otimes,a_l)$ and
$target_l=R({s_l}^\otimes,a_l)+\gamma \max_{a'} Q({{s_l}^\otimes}',a')$ such
that $({s_l}^\otimes,a_l,{{s_l}^\otimes}',R({s_l}^\otimes,a_l),{q_i}) \in
\mathcal{E}_{q_i}$.  In each epoch of LCNFQ (Algorithm~\ref{lcnfqal}), the
pattern set $\mathcal{P}_{q_i}$ is used as the input-output set to train the
neural net $B_{q_i}$.  In order to update the weights in each neural net, we
use Rprop~\cite{rprop} for its efficiency in batch learning~\cite{nfq}.  The
training schedule in the hybrid network starts from individual networks that
are associated with accepting states of the automaton.  The training
sequence goes backward until it reaches the networks that are associated to
the initial states.  By doing so, we allow the Q-value to back-propagate
through the connected networks.  In Algorithm~\ref{lcnfqal}, without loss of
generality we assume that the automaton states are ordered and hence the
back-propagation starts from $q_i=|\mathcal{Q}|$.  \rbtl{	Despite
the improvement in performance and applicability, LCNFQ cannot deal with the most
general case of MDPs, i.e., continuous state-action MDPs.  Thus, in the
following we extend LCRL towards an online learning scheme that efficiently
handles continuous state-action MDPs.}

\section{Modular Deep Actor-critic Learning}\label{sec:ddpg}

The previous section introduced LCNFQ, a model-free neural-fitted RL scheme. 
LCNFQ exploits the positive effects of generalisation in feedforward nets. 
Feedforward nets are efficient in predicting Q-values for state-action pairs
that have not been visited by interpolating between available data.  This
means that the learning algorithm requires less experience and the learning
process is thus data efficient.  However, LCNFQ is an offline learning
algorithm, i.e.,  experience gathering and learning happens separately. 
Furthermore, when dealing with the most general case of MDPs,
i.e.,~uncountably infinite-state and infinite-action, LCNFQ is of limited
use.  An obvious approach to adapt LCNFQ to continuous action domains is to
discretise the action space.  However, this has many limitations such as
loss of dynamics accuracy, and most importantly the curse of dimensionality:
the number of actions exponentially increases with the number of degrees of
freedom in the MDP.

Policy gradient methods, on the other hand, are online schemes that are
widely used in RL for MDPs with continuous action spaces.  The general idea
is to represent the policy by a parametric probability distribution
$\pi(\cdot|s,\theta^\pi)$ and then adjusting the policy parameters
$\theta^\pi$ in the direction of the greatest cumulative reward.  This
policy can of course be deterministic, but there is a crucial difference
between the stochastic and deterministic policy gradients~\cite{peters}. 
From a practical point of view, stochastic policy gradients require more
experience samples.  In this work we focus on deterministic policies, as they
are sufficient for most control problems and because deterministic policy
gradients are more efficient in terms of sample complexity.

The actor-critic architecture is a widely used method based on the policy
gradient~\cite{sutton,deeplcrl}, which consists of two interacting components: an
actor and a critic.  The actor is the parametric policy
$\pi(\cdot|s,\theta^\pi)$ \rbtl{(or $\pi(s|\theta^\pi)$ when the policy is deterministic)}, and the critic is an action-value function
$Q(s,a)$ that guides the updates of parameters $\theta^\pi$ in the direction
of the greatest cumulative reward.  The Deterministic Policy Gradient
(DPG)\nomenclature{\textbf{DPG}}{Deterministic Policy Gradient}
algorithm~\cite{DPG} introduces a parameterised deterministic function
$\pi(s|\theta^\pi)$ as the actor to represent the current policy by
deterministically mapping states to actions, where $\theta^\pi$ are the
function approximation parameters for the actor function.  A parameterised
action-value function $Q(s,a|\theta^Q)$ is the critic and is learned as
described next.

Assume that at time step $n$ the agent is at state $s_n$, takes action
$a_n$, and receives a scalar reward $R(s_n,a_n)$ as in \eqref{thereward}
with $y=1$ in $r_p$.  The action-value function update is then approximated
by parameterising $Q$ using a parameter set $\theta^Q$, i.e.,  $Q(s_n,
a_n|\theta^Q)$, and by minimising the following loss function:
\begin{equation}\label{critic}
L(\theta^Q)= \mathbb{E}^\pi_{s_n \sim \rho^\beta}[(Q(s_n, a_n|\theta^Q)-\rbtl{\varXi_n})^2],
\end{equation}
where $\rho^\beta$ is the probability distribution of state visits over $\mathcal{S}$ under any given arbitrary stochastic policy $\beta$, and $\rbtl{\varXi_n} = R(s_n,a_n)+\gamma Q(s_{n+1}, a_{n+1}|\theta^Q)$. The parameters of the actor $\pi(s|\theta^\pi)$ are updated by applying the chain rule to the expected return with respect to the actor parameters, which is approximated as follows~\cite{DPG}: 
\begin{equation*}
\begin{aligned}
&\nabla_{\theta^\pi} U^\pi(s_n) \approx \mathbb{E}_{s_n\sim p^\beta}[\nabla_{\theta^\pi} Q(s,a|\theta^Q)|_{s=s_n, a=\pi(s_n|\theta^\pi)}]  \\
&= \mathbb{E}_{s_n\sim p^\beta}[\nabla_{a} Q(s,a|\theta^Q)|_{s=s_n, a=\pi(s_n)}\nabla_{\theta^\pi}\pi(s|\theta^\pi)|_{s=s_n}].
\end{aligned}
\end{equation*}
The results in~\cite{DPG} show that this is a policy gradient, and therefore
we can apply a policy gradient algorithm on the deterministic policy.  Deep
DPG (DDPG) further extends DPG by employing a deep neural network as
function approximator and updating the network parameters via a ``soft
update'' method, similar to~\cite{deepql}, and is thoroughly explained later.

Given an LTL task and its LDGBA $\mathfrak{A}\allowbreak =
(\allowbreak\mathcal{Q}, \allowbreak q_0, \allowbreak\Sigma,
\allowbreak\mathcal{F}, \allowbreak\Delta)$, we propose a modular
architecture of $n=|\mathcal{Q}|$ separate actor, actor-target, critic and
critic-target neural networks, along with separate replay buffers.  For each
automaton state $q_i$, an actor function $\mu_{q_i}(s|\theta^{\mu_{q_i}})$
represents the current policy, where $\theta^{\mu_{q_i}}$ is the vector of
parameters of the function approximation for the actor.  The critic
$Q_{q_i}(s,a|\theta^{Q_{q_i}})$ is learned based on \eqref{critic}.

The set of neural nets acts as a global modular actor-critic deep RL
architecture, which allows the agent to jump from one sub-task to another by
just switching between the set of neural nets.  The proposed modular DDPG
algorithm is detailed in Algorithm~\ref{algor}.  Each actor-critic network
set in this algorithm is associated with its own replay buffer
$\mathcal{E}_{q_i}$, where $q_i \in \mathcal{Q}$ (lines 4 and 12).

At each time-step, actor and critic are updated by sampling a mini-batch of
size $\mathcal{M}$ uniformly from $\mathcal{E}_{q_i}$.  We only train the
actor-critic network set corresponding to the current automaton state, as
experience samples on the current automaton state have little influence on
other actor-critic networks (lines~12--17).

{Further, directly implementing the update of the critic parameters as in
\eqref{critic} is known to be potentially unstable, and as a result the
Q-update (line 14) is prone to divergence~\cite{minhd}.}
Hence, instead of directly copying the weights, the standard
DDPG~\cite{DDPG} uses ``soft'' target updates to improve learning stability. 
The target networks $Q'$ and $\mu'$ are time-delayed copies of the original
actor and critic networks that slowly track the learned networks, $Q$ and
$\mu$.
{These target actor and critic networks are used within the algorithm to
gather evidence (line~13) and subsequently to update the actor and critic
networks.}
In our algorithm, for each automaton state $q_i$ we make a copy of the actor
and the critic network, denoted by $\mu'_{q_i}(s|\theta^{\mu'_{q_i}})$ and
$Q'_{q_i}(s,a|\theta^{Q'_{q_i}})$, respectively.  The weights of both target
networks are then updated by $\theta' = \tau \theta+(1-\tau)\theta'$ with a
rate of $ \tau<1$ (line~18).

\begin{algorithm2e}[!t]
	\DontPrintSemicolon
	\SetKw{return}{return}
	\SetKwRepeat{Do}{do}{while}
	\SetKwData{conflict}{conflict}
	\SetKwData{safe}{safe}
	\SetKwData{sat}{sat}
	\SetKwData{unsafe}{unsafe}
	\SetKwData{unknown}{unknown}
	\SetKwData{true}{true}
	\SetKwInOut{Input}{input}
	\SetKwInOut{Output}{output}
	\SetKwFor{Loop}{Loop}{}{}
	\SetKw{KwNot}{not}
		\Input{LTL mission task $\varphi$, black-box model}
		\Output{actor and critic networks}
		convert the LTL property $\varphi$ to LDGBA $\mathfrak{A}=(\mathcal{Q},q_0,\Sigma, \mathcal{F}, \Delta)$\;
		randomly initialise $|\mathcal{Q}|$ actors $\mu_i(s|\theta^{\mu_i})$ and critic $Q_i(s, a|\theta^{Q_i})$ networks with weights $\theta^{\mu_i}$ and $\theta^{Q_i}$, for each $q_i \in \mathcal{Q}$, and all state-action pairs $(s,a)$\;
		initialise $|\mathcal{Q}|$ corresponding target networks $\mu'_i$ and $Q'_i$ with weights $\theta^{\mu'_i} = \theta^{\mu_i}$, $\theta^{Q'_i}= \theta^{Q_i}$\;
		initialise $|\mathcal{Q}|$ replay buffers $\mathcal{E}_{i}$\;
		\Repeat{end of trial}
		{   
			initialise $|\mathcal{Q}|$ random processes $\mathfrak{N}_i$\;
			initialise state $s_1^\otimes=(s_0,q_0)$\;
			\For{$t=1$ \textbf{to} $max\_iteration\_number$}
			{   
				choose action $a_t = \mu_{q_t}(s_t|\theta^{\mu_{q_t}})+\mathfrak{N}_{q_t}$ \;
				
				observe reward 
				$R_t$ and the new state $(s_{t+1}, q_{t+1})$\;
				
				store $((s_t,q_t), a_t, R_t, (s_{t+1}, q_{t+1}))$ in $\mathcal{E}_{q_t}$\;
				
				sample a random mini-batch of $\mathcal{M}$ transitions 
				$((s_i, q_i), a_i, R_i, (s_{i+1}, q_{i+1}))$ from $\mathcal{E}_{q_t}$\;
				
				set $\rbtl{\varXi_i} = R_i + \gamma Q_{q_{i+1}}'(s_{i+1}, \mu'_{q_{i+1}}(s_{i+1}|\theta^{\mu'_{q_{i+1}}})|\theta^{Q'_{q_{i+1}}})$\;
				
				update critic $Q_{q_t}$ and $\theta^{Q_{q_t}}$ by minimising the loss:
				$L =1/|\mathcal{M}|\sum_{i}(\rbtl{\varXi_i}-Q_{q_t}(s_i, a_i|\theta^{Q_{q_t}}))^2$\;
				
				update the actor policy $\mu_{q_t}$ and $\theta^{\mu_{q_t}}$ by maximising the sampled 
				policy gradient:\;
				$\nabla_{\theta^{\mu_{q_t}}} U^{\mu_{q_t}} \approx 1/|\mathcal{M}| \sum_{i}[\nabla_{a} Q_{q_t}(s,a|\theta^{Q_{q_t}})|_{s=s_i, a=\mu_{q_t}(s_i|\theta^{\mu_{q_t}})}$\\ ~~$\nabla_{\theta^{\mu_{q_t}}}\mu_{q_t}(s|\theta^{\mu_{q_t}})|_{s=s_i}]$
				
				update the target networks:
				$\theta^{Q'_{q_t}} \leftarrow \tau\theta^{Q_{q_t}} + (1-\tau)\theta^{Q'_{q_t}}$
				$\theta^{\mu'_{q_t}} \leftarrow \tau\theta^{\mu^{q_t}} + (1-\tau)\theta^{\mu'_{q_t}}$\;
			}
		}
	\caption{Modular DDPG}
	\label{algor}
\end{algorithm2e}

\begin{figure}[!t]\centering
	\subfloat[]{{\includegraphics[width=0.52\columnwidth]{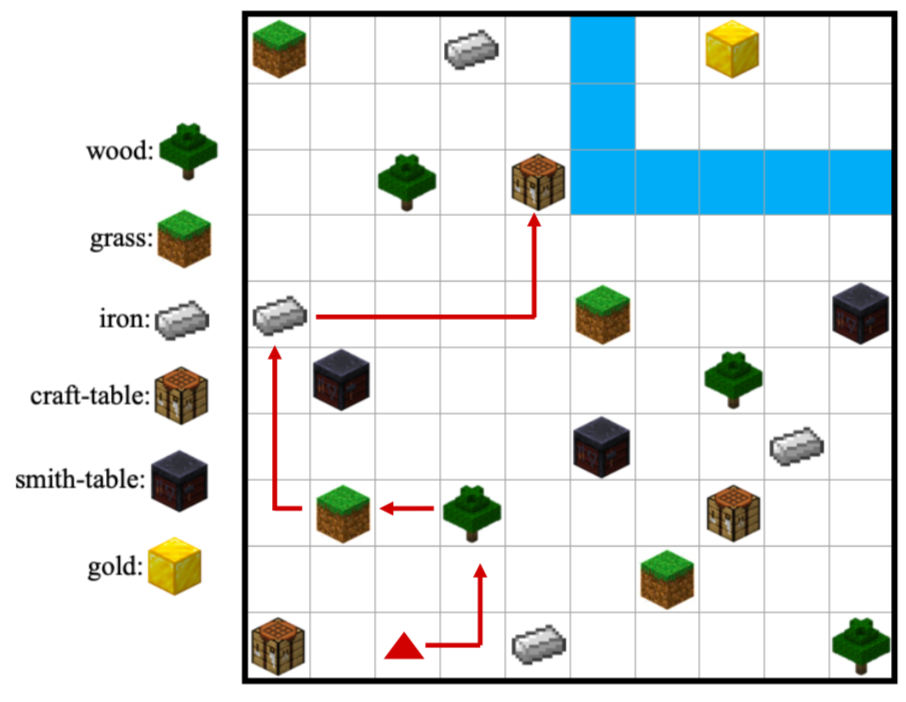} }}%
	\subfloat[]{{\includegraphics[width=0.43\columnwidth]{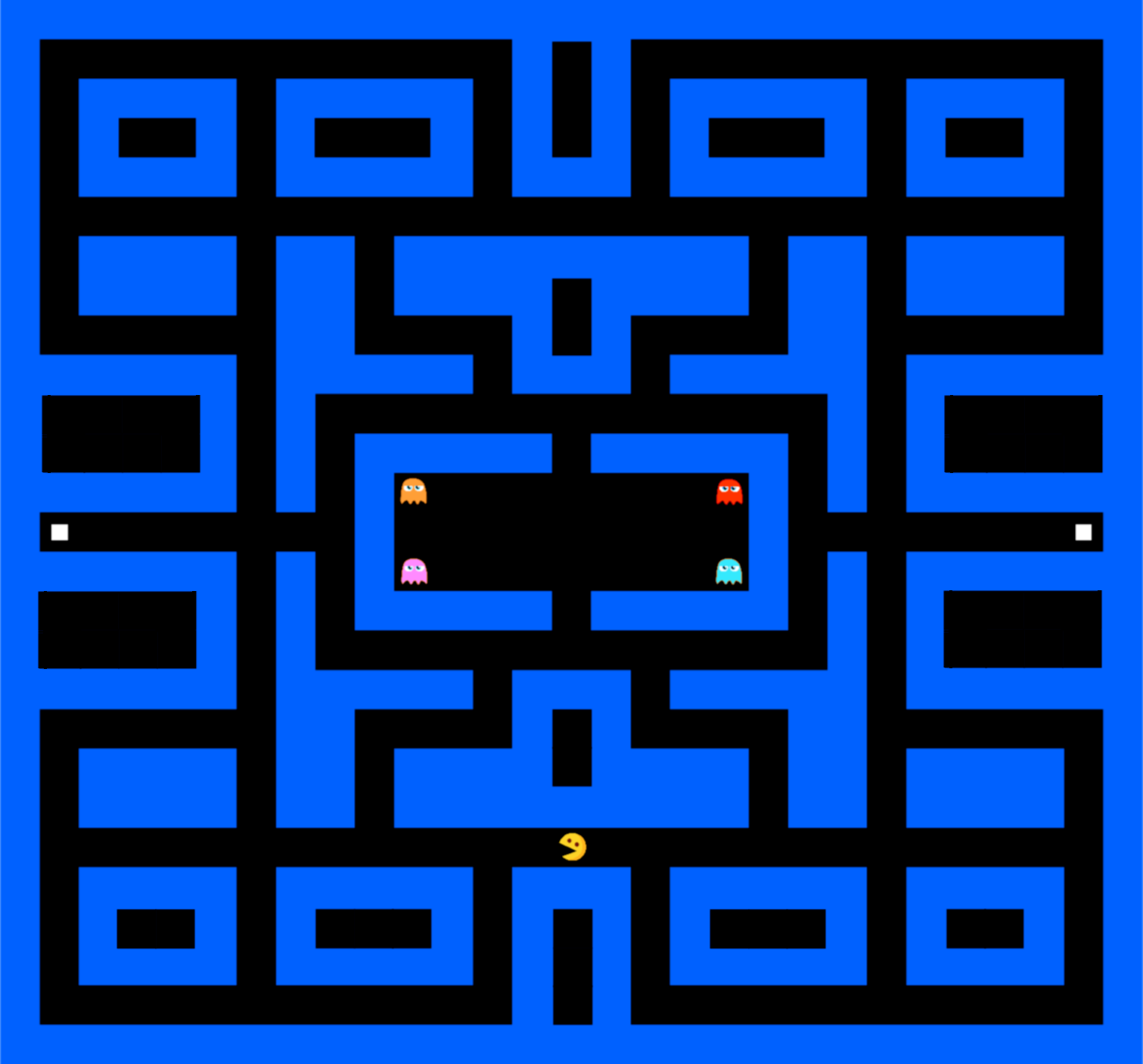} }}%
	\caption{(a) sample run by LCRL policy learnt for Task 3. (b)
	$\mathtt{pacman\text{-}lrg}$---the square on the left is labelled as food 1 ($ f_1 $) and the one on the right as food 2 ($ f_2 $), the state of being caught by a ghost is labelled as ($ g $) and the rest of the state space is neutral ($ n $).}
	\label{minecraft_task_3_path}
\end{figure}
\begin{figure}[!t]
	\centering
	\subfloat[]{{\includegraphics[width=0.45\columnwidth]{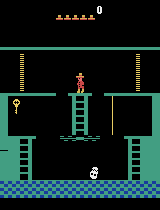} }}%
	\subfloat[]{{\includegraphics[width=0.45\columnwidth]{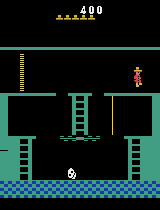} }}%
	\caption{(a) \texttt{montezuma} (Montezuma's Revenge) initial frame (b) agent successfully unlocks a door.}
	\label{montezuma_success}
\end{figure}

\begin{figure}[!t]
	\centering
	\subfloat[]{{\includegraphics[width=0.4\columnwidth]{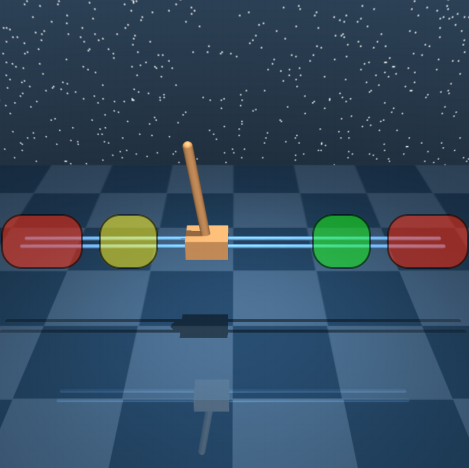} }}%
	\subfloat[]{{\includegraphics[width=0.58\columnwidth]{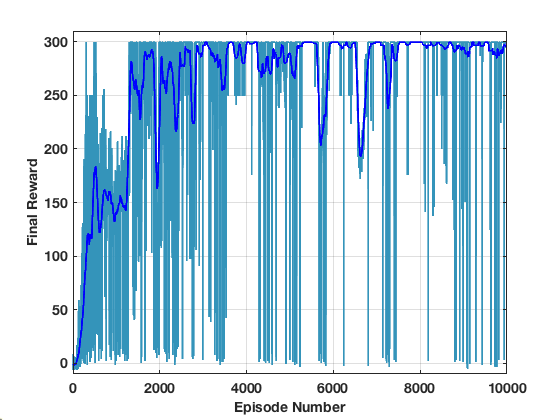} }}%
	\caption{(a) $\mathtt{cart\text{-}pole}$ experiment~\cite{tassa2018deepmind}; (b) learning curves (dark blue) obtained averaging over 10 randomly initialised experiment, shaded areas (light blue) represent envelopes of 10 generated learning curves.}%
	\label{MDPcart}%
\end{figure}

\section{Experimental Results} \label{case study}

We have tested LCRL on a number of planning experiments that require policy
synthesis for a given temporal specification, both when the state and action
spaces are finite, and when they are continuous. \aij{Despite dealing with an unknown MDP, we have observed that LCRL results are comparable to those of model-checking methods whenever available, as will be discussed in detail in this section}. All the experiments are run on a
standard PC, with an Intel Core i5 CPU at 2.5\,GHz $\times$ 4 and with 20~GB
of RAM.  

\bgroup
\def\arraystretch{0.9}
\begin{table}[!ht]
	\centering
	\caption{\rbtl{Learning results with LCRL. MDP state and action space cardinalities are $|\mathcal{S}|$ and $|\mathcal{A}|$, the number of automaton states in LDBA is denoted by $|\mathcal{Q}|$, 
			the optimal action value function in the initial state is denoted by ``LCRL $\max_a Q(s_0,a)$'', which represents the LCRL estimation of the maximum satisfaction probability. For each experiment, the reported result includes the mean and the standard error of ten learning trials with LCRL. 
			This probability is also calculated by the PRISM model checker \cite{prism}, whenever the MDP model can be processed by PRISM, and accordingly it is reported in the column ``\texttt{max sat. prob. at $s_0$}''. 
			The closer ``LCRL $\max_a Q(s_0,a)$'' and ``\texttt{max sat. prob. at $s_0$}'' the better. Note that for continuous state-action MDPs the maximum satisfaction probability cannot be precisely computed by model checking tools, unless abstraction approximation techniques (outside of the scope of this work, cf. \cite{lavaei2021automated}) are applied, hence ``n/a''. Furthermore, if the MDP state (or action) space is large enough, e.g. \texttt{pacman}, the process (either learning or model-checking) times out, i.e. ``t/o''. The column ``\texttt{episode\_num}'' presents the episode number in which LCRL converged, using LDBA and also DRA as the underlying automaton. 
			The rest of the columns provide the values
			of the hyper-parameters across the different benchmarks.}}
	\label{tab:results}
	\vspace{2mm}
	{
		\resizebox{1\textwidth}{!}{%
			\begin{tabular}{|l|c|c|c|c|c|c|c|c|c|c|c|c|}
				\hline
				\multirow{ 2}{*}{~\texttt{experiment}} \multirow{ 2}{*}&{\texttt{MDP}} & \texttt{LDBA} & LCRL $\max_a$& \texttt{max sat.} & \multirow{ 2}{*}{\texttt{alg.}} &{\texttt{episode\_}} & \texttt{iteration\_}& \texttt{discount\_}& \texttt{learning\_}&\texttt{wall\_clock}\\
				& ~{$|\mathcal{S}|, |\mathcal{A}|$}~ &~{$|\mathcal{Q}|$}~ & $Q(s_0,a)$ & \texttt{prob. at} $s_0$ &&\texttt{num} {\tiny (LDBA/DRA)}& {\texttt{num\_max}} & \texttt{factor}{\Large ${}^*$}& \texttt{rate{\large ${}^\dag$}}&\texttt{time{${}^\bigstar$(min)}}\\ \hline
				\texttt{minecraft-t1}  &  100, 5 & 3 &   0.991 $\pm$ 0.009& 1 & \texttt{`ql'} & 500/1000 & 4000 & 0.95 & 0.9 & 0.1 \\
				\texttt{minecraft-t2}  &  100, 5 & 3 &   0.991 $\pm$ 0.009 & 1 & \texttt{`ql'} & 500/1000 & 4000 & 0.95 & 0.9 & 0.1\\
				\texttt{minecraft-t3}  &  100, 5 & 5 &   0.993 $\pm$ 0.007 & 1 & \texttt{`ql'} & 1500/6000 & 4000 & 0.95 & 0.9 & 0.25\\
				\texttt{minecraft-t4}  &  100, 5 & 3 &   0.991 $\pm$ 0.009 & 1 & \texttt{`ql'} & 500/1000 & 4000 & 0.95 & 0.9 & 0.1\\
				\texttt{minecraft-t5}  &  100, 5 & 3 &   0.995 $\pm$ 0.005 & 1 & \texttt{`ql'} & 500/1000 & 4000 & 0.95 & 0.9 & 0.1\\
				\texttt{minecraft-t6}  &  100, 5 & 4 &   0.995 $\pm$ 0.005 & 1 & \texttt{`ql'} & 1500/4200 & 4000 & 0.95 & 0.9 & 0.25\\
				\texttt{minecraft-t7}  &  100, 5 & 5 &   0.993 $\pm$ 0.007 & 1 & \texttt{`ql'} & 1500/6000 & 4000 & 0.95 & 0.9 & 0.5\\
				\texttt{mars-rover-1}  &$\infty$, 5& 3 & 0.991 $\pm$ 0.002& n/a & \texttt{`nfq'} & 50/170 & 3000 & 0.9 & 0.01 & 550\\
				\texttt{mars-rover-2}  &$\infty$, 5& 3 & 0.992 $\pm$ 0.006 & n/a & \texttt{`nfq'} & 50/150 & 3000 & 0.9 & 0.01 & 540\\
				\texttt{mars-rover-3}  &$\infty$, $\infty$& 3 & n/a & n/a & \texttt{`ddpg'} & 1000/3800 & 18000 & 0.99 & 0.05 & 14\\
				\texttt{mars-rover-4}  &$\infty$, $\infty$& 3 & n/a & n/a & \texttt{`ddpg'} & 1000/3000 & 18000 & 0.99 & 0.05 & 12\\
				\texttt{mars-rover-5}  &$\infty$, $\infty$& 13 & n/a & n/a & \texttt{`ddpg'} & 17e3/ t/0 & 25000 & 0.9 & 0.003 & 300\\
				\texttt{cart-pole}  &$\infty$, $\infty$& 4 & n/a & n/a & \texttt{`ddpg'} & 100/420 & 10000 & 0.99 & 0.02 & 1\\
				\texttt{montezuma}  &$\infty$, 18& 7 & n/a & n/a & \texttt{`dqn'} & 400e3/ t/o & 150e3 & 0.99 & 0.00025 & $>$4000\\
				\texttt{robot-surve}   &   25, 4 & 3 &  0.994 $\pm$ 0.006 & 1 & \texttt{`ql'} & 500/1000 & 1000 & 0.95 & 0.9 & 0.1\\
				\texttt{slp-easy-sml}  &  120, 4 & 2 &  0.974 $\pm$ 0.026 & 1 & \texttt{`ql'} & 300/500 & 1000 & 0.99 & 0.9 & 0.1\\
				\texttt{slp-easy-med}  &  400, 4 & 2 & 0.990 $\pm$ 0.010 & 1 & \texttt{`ql'} & 1500/2700 & 1000 & 0.99 & 0.9 & 0.25\\
				\texttt{slp-easy-lrg}  & 1600, 4 & 2 & 0.970 $\pm$ 0.030 & 1 & \texttt{`ql'} & 2000/3500 & 1000 & 0.99 & 0.9 & 2\\
				\texttt{slp-hard-sml}  &  120, 4 & 5 &  0.947 $\pm$ 0.039 & 1 & \texttt{`ql'} & 500/1800 & 1000 & 0.99 & 0.9 & 1\\
				\texttt{slp-hard-med}  &  400, 4 & 5 & 0.989 $\pm$ 0.010 & 1 & \texttt{`ql'} & 4000/9000 & 2100 & 0.99 & 0.9 & 5\\
				\texttt{slp-hard-lrg}  & 1600, 4 & 5 & 0.980 $\pm$ 0.016 & 1 & \texttt{`ql'} & 6000/15000& 3500 & 0.99 & 0.9 & 9\\
				\texttt{frozen-lake-1} &  120, 4 & 3 &  0.949 $\pm$ 0.050 & 0.9983 & \texttt{`ql'} & 400/800 & 2000 & 0.99 & 0.9 & 0.1\\
				\texttt{frozen-lake-2} &  400, 4 & 3 & 0.971 $\pm$ 0.024 & 0.9982 & \texttt{`ql'} & 2000/4100 & 2000 & 0.99 & 0.9 & 0.5\\
				\texttt{frozen-lake-3} & 1600, 4 & 3 &  0.969 $\pm$ 0.019 & 0.9720 & \texttt{`ql'} & 5000/9400 & 4000 & 0.99 & 0.9 & 1\\
				\texttt{frozen-lake-4} &  120, 4 & 6 &  0.846 $\pm$ 0.135 & 0.9728 & \texttt{`ql'} & 2000/9000 & 2000 & 0.99 & 0.9 & 1\\
				\texttt{frozen-lake-5} &  400, 4 & 6 & 0.735 $\pm$ 0.235 & 0.9722 & \texttt{`ql'} & 7000/ t/o & 4000 & 0.99 & 0.9 & 2.5\\
				\texttt{frozen-lake-6} & 1600, 4 & 6 & 0.947 $\pm$ 0.011 & 0.9467 & \texttt{`ql'} & 5000/ t/o & 5000 & 0.99 & 0.9 & 9\\
				\texttt{pacman-sml} & 729,000, 5 & 6 & 0.290 $\pm$ 0.035 & t/o${}^\ddag$ & \texttt{`ql'} & 80e3/400e3 & 4000 & 0.95 & 0.9 & 1600\\
				\texttt{pacman-lrg} & 4,251,000, 5 & 6 & 0.282 $\pm$ 0.049 & t/o${}^\ddag$ & \texttt{`ql'} & 180e3/ t/o & 4000 & 0.95 & 0.9 & 3700\\
				\hline              
		\end{tabular}}
	}
	{\\\vspace{1mm} * coefficient $\eta$ in \eqref{gamma} ~ $\dag$ learning rate $\mu$  ~ $\ddag$ timed out due to the state space size~ ${}^\bigstar$ on a machine running macOS~11.6.5 with Intel Core i5 CPU at 2.5~GHz and with 20~GB of RAM}
	\vspace{-4mm}
\end{table}
\egroup

The experiments are listed in Table~\ref{tab:results}.  For each MDP, the
state space cardinality $|\mathcal{S}|$ and the action space cardinality
$|\mathcal{A}|$ are given, and for each LTL property the number of automaton
states $|\mathcal{Q}|$ is reported.  \aij{For each MDP-automaton product pair, 
Table~\ref{tab:results} presents the maximum probability of satisfaction of
the LTL objective at the initial state when using the strategy synthesised
by LCRL.  As a reference, we also compute the maximum satisfaction
probability using the PRISM model checker~\cite{prism} and report it in the table,
whenever computationally feasible (the model checker has scalability limits, as expected, that prevent completion on two of the benchmarks).}  The
remaining columns give the values of the hyper-parameters.  All the reported
results are the averages of ten learning trials done with LCRL.

The \texttt{minecraft} environment, adopted from~\cite{pol-sketch}, requires
solving challenging low-level control tasks
($\mathtt{minecraft\text{-}tX}$), and features many sequential high-level
goals.  For instance, in $\mathtt{minecraft\text{-}t3}$
(Fig.~\ref{minecraft_task_3_path}.a) the agent has to collect three items
sequentially, and to reach a final checkpoint, which is encoded as the
following LTL formula:
\begin{equation*}
\begin{aligned}
&\Diamond(\mathit{wood}\wedge\Diamond (\mathit{grass} \wedge\Diamond(\mathit{iron} \wedge\Diamond(\mathit{craft\_table})))) 
\end{aligned}
\label{minecraft_task_3_spec}
\end{equation*}

The $\mathtt{mars\text{-}rover}$ problems are realistic benchmarks taken
from~\cite{lcnfq,hasanbeig2020deep}, and the models feature uncountably infinite (continuous)
state and action spaces. Fig.~\ref{example_automata}.a gives the constructed automaton for $\mathtt{mars\text{-}rover\text{-}5}$. The $\mathtt{cart\text{-}pole}$ experiment (Fig.~\ref{MDPcart}) is explained
in~\cite{tassa2018deepmind,hasanbeig2020deep,cai2021}, and the property of interest is expressed by the LTL formula 
\begin{equation*}\label{c_ltl}
\square\lozenge y \wedge \square\lozenge g \wedge \square \neg u.
\end{equation*} 
\hos{The pole starts upright, and the goal is to prevent the pendulum from
falling over ($\square \neg u$), and to move the cart between the yellow
($y$) and green ($g$) regions ($\square\lozenge y \wedge \square\lozenge
g$), while avoiding the red (unsafe) parts of the track ($\square \neg u$).}

\begin{figure}[!t]
	\centering
	\subfloat[]{
		\scalebox{0.5}{
			\begin{tikzpicture}[shorten >=1pt,node distance=2cm,on grid,auto] 
				\node[state,initial] (q_1)   {$q_1$}; 
				\node[state] (q_2) [right=of q_1] {$q_2$};
				\node[state] (q_3) [right=of q_2] {$q_3$};
				\node[state] (q_4) [right=of q_3] {$q_4$};
				\node[state] (q_5) [right=of q_4] {$q_5$};
				\node[state] (q_6) [right=of q_5] {$q_6$};
				\node[state] (q_7) [right=of q_6] {$q_7$};
				\node[state] (q_8) [right=of q_7] {$q_8$};
				\node[state] (q_9) [right=of q_8] {$q_9$};
				\node[state] (q_10) [right=of q_9] {$q_{10}$};
				\node[state] (q_11) [right=of q_10] {$q_{11}$};
				\node[state] (q_13) [below= 4cm of q_6] {$q_{13}$}; 
				\node[state,accepting] (q_12) [right=of q_11] {$q_{12}$}; 
				\path[->] 
				(q_1) edge [loop above] node {$\neg t_1$} ()   	
				(q_1) edge  node {$t_1$} (q_2)
				(q_2) edge [loop above] node {$\neg t_2$} ()
				(q_2) edge node {$t_2$} (q_3)
				(q_3) edge node {$t_3$} (q_4)
				(q_4) edge node {$t_4$} (q_5)
				(q_5) edge node {$t_5$} (q_6)
				(q_6) edge node {$t_6$} (q_7)
				(q_7) edge node {$t_7$} (q_8)
				(q_8) edge node {$t_8$} (q_9)
				(q_9) edge node {$t_9$} (q_10)
				(q_10) edge node {$t_{10}$} (q_11)
				(q_11) edge node {$t_{11}$} (q_12)
				(q_3) edge [loop above] node {$\neg t_3$} () 
				(q_4) edge [loop above] node {$\neg t_4$} ()
				(q_5) edge [loop above] node {$\neg t_5$} ()
				(q_6) edge [loop above] node {$\neg t_6$} ()
				(q_7) edge [loop above] node {$\neg t_7$} ()
				(q_8) edge [loop above] node {$\neg t_8$} ()
				(q_9) edge [loop above] node {$\neg t_9$} ()
				(q_10) edge [loop above] node {$\neg t_{10}$} ()
				(q_11) edge [loop above] node {$\neg t_{11}$} ()
				
				(q_1) edge [bend right] node {$u$} (q_13)
				(q_2) edge [bend right] node {$u$} (q_13)
				(q_3) edge [bend right] node {$u$} (q_13)
				(q_4) edge [bend right] node {$u$} (q_13)
				(q_5) edge [bend right] node {$u$} (q_13)
				(q_6) edge node {$u$} (q_13)
				(q_7) edge [bend left] node {$u$} (q_13)
				(q_8) edge [bend left] node {$u$} (q_13)
				(q_9) edge [bend left] node {$u$} (q_13)
				(q_10) edge [bend left] node {$u$} (q_13)
				(q_11) edge [bend left] node {$u$} (q_13)
				(q_12) edge [loop above] node {$t_{12}$} ()
				(q_13) edge [loop below] node {$\mathit{True}$} ();
	\end{tikzpicture}} 
	}
\\*
	\subfloat[]{
		\scalebox{0.7}{
			\begin{tikzpicture}[shorten >=1pt,node distance=2.3cm,on grid,auto] 
				\node[state,initial] (q_1) {$q_0$}; 
				\node[state] (q_2) [above right=of q_1] {$q_1$}; 
				\node[state] (q_3) [below right=of q_1] {$q_2$}; 
				\node[state] (q_5) [right=of q_1] {$q_4$}; 
				\node[state,accepting] (q_4) [right=of q_5] {$q_3$}; 
				\path[->] 
				(q_1) edge [loop below] node {$n$} ()   	
				(q_1) edge [bend left=15] node {$f_1$} (q_2)
				(q_1) edge [bend right=15] node {$f_2$} (q_3)
				(q_2) edge [loop above] node {$n \vee f_1$} ()
				(q_2) edge [bend right=-15] node {$f_2$} (q_4)
				(q_3) edge [loop below] node {$n \vee f_2$} ()
				(q_3) edge [bend right=15] node {$f_1$} (q_4) 
				(q_2) edge node {$g$} (q_5)
				(q_3) edge node {$g$} (q_5)
				(q_1) edge node {$g$} (q_5)
				(q_5) edge [loop right] node {$\mathit{true}$} ()
				(q_4) edge [loop right] node {$\mathit{true}$} ();
		\end{tikzpicture}}
	}%
	\subfloat[]{
		\scalebox{0.7}{
			\begin{tikzpicture}[shorten >=1pt,node distance=2.8cm,on grid,auto] 
				\node[state,initial,accepting] (q_0)   {$q_0$}; 
				\node[state] (q_1) [above right=of q_0] {$q_1$}; 
				\node[state] (q_2) [below right=of q_1] {$q_2$}; 
				\path[->] 
				(q_0) edge  node {$\neg B \wedge \neg C$} (q_1)
				(q_0) edge node {$B \wedge \neg A \wedge \neg C$} (q_2)
				(q_2) edge [bend left=45] node {$A \wedge \neg C$} (q_0)
				(q_1) edge  node {$B \wedge \neg A \wedge \neg C$} (q_2)
				(q_2) edge [loop right] node {$\neg A \wedge \neg C$} ()
				(q_1) edge [loop above] node {$\neg B \wedge \neg C$} ();    
		\end{tikzpicture}}
	}%
	\caption{\aij{LDBA for the specification in (a) \texttt{mars-rover-5}; (b) $\mathtt{pacman\text{-}sml}$ and
			$\mathtt{pacman\text{-}lrg}$; and (c) \texttt{robot-surve}.}}%
	\label{example_automata}%
\end{figure}

Atari 2600 Montezuma's Revenge $\mathtt{montezuma}$ is an infamously hard
exploration problem.  \hos{The task in $\mathtt{montezuma}$
(Fig~\ref{montezuma_success}.a) is to climb down to the floor from the top
of the middle ladder, jump over the skull, fetch the key and move back up
to open either doors.} The example $\mathtt{robot\text{-}surve}$ is adopted
from~\cite{dorsa}, and the task is to visit two regions ($A$ and $B$) in
sequence, while avoiding multiple obstacles ($C$) on the way: $
\square\lozenge A \wedge \square\lozenge B \wedge \square \neg C$.
The LDGBA expressing this LTL formula is presented in Fig.~\ref{example_automata}.c. Models $\mathtt{slp\text{-}easy}$ and $\mathtt{slp\text{-}hard}$ are
inspired by the widely used noisy MDPs in~\cite[Chapter~6]{sutton}: the goal
in $\mathtt{slp\text{-}easy}$ is to reach a particular region of the MDP,
whereas the goal in $\mathtt{slp\text{-}hard}$ is to visit four distinct
regions sequentially in a given order.  The $\mathtt{frozen\text{-}lake}$
benchmarks are similar: the first three are reachability problems, whereas
the last three require sequential visits of four regions in the presence of
unsafe regions as well.  The $\mathtt{frozen\text{-}lake}$ MDPs are adopted
from the OpenAI Gym~\cite{gym}.

Finally, $\mathtt{pacman\text{-}sml}$ and
$\mathtt{pacman\text{-}lrg}$ are inspired by the well-known Atari game
Pacman, and are initialised in a tricky configuration
($\mathtt{pacman\text{-}lrg}$ as in Fig.~\ref{minecraft_task_3_path}.b)
likely to lead to getting caught: to win the game the agent has to collect
all available tokens without being caught by moving ghosts.  Formally, the
agent is required to choose between one of the two available foods and then
find the other one ($ \lozenge [ (f_1 \wedge \lozenge f_2) \vee (f_2 \wedge
\lozenge f_1)] $) while avoiding the ghosts ($ \square \neg g$). We
feed a conjunction of these associations to the agent by using the following
LTL formula$
\label{pacman_p}
\lozenge [ (f_1 \wedge \lozenge f_2) \vee (f_2 \wedge \lozenge f_1)]  \wedge \square \neg g$.
The LDGBA expressing this LTL formula is presented in Fig.~\ref{example_automata}.b. Standard QL failed to find a policy with satisfying traces. \aij{Similar to LCRL, the reward function in standard QL is positive if the agent manages to achieve the specified task, and zero otherwise. However, one major difference is that the state space in standard QL is not enriched with the automaton, and hence any generated policy is memoryless.} 

\section{Related Work}
\label{sec:related}

The goal of control (policy) synthesis in finite-state/action MDPs, under
temporal logic specifications, has been considered in numerous works. 
In~\cite{wolf}, the property of interest is expressed in LTL and converted
to a DRA to synthesise a robust control policy.  Specifically, a product MDP
is constructed with the resulting DRA and a modified DP is applied to the
product MDP, maximising the worst-case probability of satisfying the
specification over all transition probabilities.  However,~\cite{wolf}
assumes that the MDP is known a-priori.  This assumption is relaxed
in~\cite{topku} and transition probabilities in the given MDP model are
considered to be unknown.  Instead, a Probably Approximately Correct MDP
(PAC MDP) is constructed and further multiplied by the logical property
after conversion to a DRA.  The overall goal is to calculate a
finite-horizon $T$-step value function for each state such that the obtained
value is within an error bound from the probability of satisfying the given
LTL property.

The goal of maximising the probability of satisfying unbounded-time
reachability properties, when the MDP transition probabilities are unknown,
is investigated in~\cite{brazdil}.  The policy generation relies on
approximate DP, which requires a mechanism to approximate these
probabilities (much like the PAC MDP above), and the quality of the
generated policy critically depends on the accuracy of this approximation,
which might require a large number of simulation runs.
Furthermore, the algorithm in~\cite{brazdil} assumes prior knowledge about
the smallest transition probability in the MDP.  Using an LTL-to-DRA
conversion, the algorithm given in~\cite{brazdil} can be extended to the
problem of control synthesis for LTL specifications, at the expense of a
double-exponential blow-up of the obtained automaton.

Much in the same direction, if there exists a policy whose traces satisfy
the property with probability one, \cite{dorsa} employs a learning-based
approach to generate one such policy.
As for~\cite{brazdil}, the algorithm in~\cite{dorsa} hinges on approximating
the transition probabilities, which affects precision and scalability of the
approach.

Our work on model-free RL for LTL has been first introduced in~\cite{arxiv}. 
Since then, growing research has been devoted to model-free RL with
different kinds of automata~\cite{cai2020}, including limit-deterministic
B\"uchi automata~\cite{hahn,bozkurt}, where different forms of reward
schemes have been
examined~\cite{lennartson2020,cai2021soft,mingyu2021optimal,kantaros2022accelerated,kalluraya2023resilient}. 
Specifically,~\cite{hahn} employs a limit-deterministic B\"uchi automata
based on a Non-deterministic B\"uchi Automaton (NBA) that is structurally
different than~\cite{sickert}'s LDGBA, used in this work.  Bozkurt and
others~\cite{bozkurt} extend upon the NBA-based automata, by proposing an
interleaving reward and discounting scheme, and~\cite{cai2021} employs
policy gradient to tackle MDPs with high-dimensional state/action spaces.

There has been significant interest also in specifying tasks in RL via
sub-fragments of LTL~\cite{toro,camacho,camacho2,lavaei2020formal} and other
forms of temporal logic~\cite{kapoor2020model,wang2020statistically,bansal2022synthesis}.  The
problem of synthesising a policy that satisfies a temporal logic
specification and that at the same time optimises a performance criterion is
considered in~\cite{lesser,belta2,uppaal,game}.  In~\cite{scltl}, scLTL is
proposed for mission specifications, which results in Deterministic Finite
Automata (DFAs).  A product MDP is then constructed and a linear programming
solver is used to find optimal policies.~\cite{deepsynth,memarian2020active}
synthesise DFAs on-the-fly in the context of deep RL and inverse RL, to
infer an scLTL property.  Conversely, Mealy machines are inferred
in~\cite{rens2020learning} to represent a temporal non-Markovian task in
Monte Carlo tree search.  PCTL specifications are investigated
in~\cite{morteza}, where a linear optimisation solution is used to
synthesise a control policy.  In~\cite{pmc}, an automated method is proposed
to verify and repair the policies that are generated by RL with respect to a
PCTL formula - the key engine runs by feeding the Markov chain induced by
the policy to a probabilistic model checker.  In~\cite{andersson}, practical
challenges of RL are addressed by letting the agent plan ahead in real time
using constrained optimisation.

In~\cite{game}, the problem is separated into two sub-problems: extracting a
(maximally) permissive strategy for the agent and then quantifying the
performance as a reward function and computing an optimal strategy for the
agent within the operating envelope allowed by the permissive strategy. 
Similarly,~\cite{nils} first computes safe, permissive strategies with
respect to a reachability property.  Then, under these constrained
strategies, RL is applied to synthesise a policy that minimises an expected
cost.  The concept of shielding is employed in~\cite{shield} to synthesise a
policy that ensures that the agent remains safe during and after learning for a fully-deterministic reactive system.  This approach is closely
related to teacher-guided RL~\cite{teacher}.
In order to express the temporal specification,~\cite{shield} uses DFAs and
then translates the problem into a safety game.  The game is played between
the environment and the agent, where in every state of the game the
environment chooses an input, and then the agent selects an output.  The
game is won by the agent if only safe states are visited during the play. 
However, the generated policy always needs the shield to be online, as the
shield maps every unsafe action to a safe action.  The work
in~\cite{shield2} extends~\cite{shield} to probabilistic systems modelled as
MDPs with adversarial uncontrollable agents.  The general assumption
in~\cite{shield2} is that the controllable agent acquires full observations
over the MDP and the adversarial agent: unlike the proposed method in this
work, the RL scheme used in~\cite{shield2} is model-based and requires the
agent to first build a model of the stochastic MDP.  The concept of a
bounded-prescience shield is proposed in~\cite{bounded_shield} for analysing
and ensuring the safety of deep RL agents in Atari games. 
\cite{fulton2,fulton,fulton4,fulton5} address safety-critical settings in
the context of cyber-physical systems, where the agent has to deal with a
heterogeneous set of models in model-based RL.~\cite{fulton} first generates
a set of feasible models given an initial model and data on runs of the
system.  With such a set of feasible models, the agent has to learn how to
safely identify which model is the most accurate one.~\cite{fulton3} further
employs differential dynamic logic~\cite{ddl}, a first-order multimodal
logic for specifying and proving properties of hybrid models.

Safe RL is an active area of research whose focus is on the efficient
implementation of safety properties, and is mostly based on reward
engineering~\cite{garcia,belzner2020synthesizing}.  Our proposed method is
related to work on safe RL, but cannot simply be reduced to it, due to its
generality and to its inherent structural differences.  By focusing on the
safety fragment of LTL, the proposed scheme does not require the reward
function to be handcrafted.  The reward function is automatically shaped by
exploiting the structure of the LDGBA and its generalised B\"uchi acceptance
condition.  However, for the safety fragment of LTL the proposed automatic
reward shaping can be seen as a way of ``modifying the optimisation
criterion,'' as in~\cite{garcia}.  Additionally, we would like to emphasise
that our work cannot be considered a Constrained MDP (CMDP) method, as the
LTL satisfaction is encoded in the expected return itself, while in CMDP
algorithms the original objective is separated from the constraint.  In a
nutshell, the proposed method inherits reward engineering aspects that are
standard in safe RL, however at the same time it infuses notions from formal
methods that allow guiding exploration and certifying its outcomes. 
\hos{Note that safe RL in general is not equivalent to ensuring agent safety
during learning.  Learning-while-being-safe by itself is a
semantically-different research area that has attracted considerable
attention recently, e.g.,~\cite{polymenakos2017safe, cautiousRL,
belzner2020synthesizing, grbic2020safe, turchetta2020safe,
turchetta2019safe,cai2021safety,salamati2021data,mittarisk}}.

A relevant body of work had been done on both finite- and
continuous-state-action MDPs, when the MDP model is fully known~\cite{lavaei2021automated}. 
Probabilistic reachability over a finite horizon for hybrid
continuous-state-action MDPs is investigated
in~\cite{reachability_in_hybrid}, where a DP-based algorithm is employed to
produce safe policies.  DFAs have been employed in~\cite{tkachev2} to find
an optimal policy for infinite-horizon probabilistic reachability problems. 
FAUST$^2$~\cite{faust}, StocHy~\cite{stochy}, and
AMYTISS~\cite{lavaei2020amytiss} deal with uncountable-state MDPs by
generating a discrete-state abstraction based on the knowledge of the MDP
model.  Using probabilistic bi-simulation~\cite{bisim} showed that
abstraction-based model checking can be effectively employed to generate
control policies in continuous-state/action MDPs.  Bounded LTL is proposed
in~\cite{belta} as the specification language, and a policy search method is
used for synthesis.  Automatic control approaches are also studied to
deal with partially infeasible LTL specifications~\cite{cai2020receding,
cai2020optimal}.

Statistical Model Checking (SMC)\nomenclature{\textbf{SMC}}{Statistical
Model Checking} techniques have also been studied for policy synthesis in
MDPs, however they are not well suited to models that exhibit
non-determinism.  This is due to the fact that SMC techniques often rely on
generation of random paths, which are not well-defined for an MDP with
non-determinism~\cite{smc_mdp_0,smc_mdp_0_1}.  Some~SMC approaches proposed
to resolve the MDP non-determinism by using uniform
distributions~\cite{smc_mdp_1,smc_mdp_2} and others proposed to consider all
possible strategies~\cite{smc_mdp_3,zuliani} and produced policies that are
close to an optimal one.  Unlike RL, which improves its exploration policy
during learning, a constant random policy is expected to waste time and
computational resources to generate sample traces.  Also, a trace is
``only'' used to reinforce each state-action pair visited by the associated
path if the trace satisfies the property of interest.  This is quite similar
to Monte Carlo methods rather than RL or DP.  For this reason, SMC methods
are not expected to scale as well as RL.  Further, sampling and checking of
traces needs to be computationally feasible: SMC techniques are effective
with finite-horizon LTL properties, as opposed to the focus of this work on
infinite-horizon properties and full LTL.  The efforts on statistical
model-checking of unbounded properties is limited to a few
specifications~\cite{smc_mdp_4}.  However, there have been some recent
developments, e.g.,~\cite{zuliani}, that leverage RL to reduce the randomness
in the policy.

\smallskip

\rbtl{This article summarises and extends material presented in
earlier conference
publications~\cite{cai2021,lcnfq,cautiousRL,plmdp,hasanbeig2020deep,overlay,hasanbeig2022lcrl},
whilst presenting it in a more comprehensive manner and with a unique set of
experiments.  More specifically, in this article we present and expand
rigorous theoretical guarantees, upon which our earlier
work~\cite{cai2021,overlay} was established.  This includes new definitions,
propositions, and theorems on automata and RL theory along with complexity
analysis.  Furthermore, we present new experiments that are significantly
more complicated and sophisticated than those presented
in~\cite{cai2021,overlay}---an example is the well-known Atari game
Montezuma's Revenge which is still a hard problem for modern AI solutions. 
These new experimental results are a significant improvement over the
results presented in~\cite{lcnfq,cautiousRL,plmdp,hasanbeig2020deep}, and
further support the claims on performance, correctness, and sample
efficiency~\cite{hasanbeig2020safe}.}

\section{Conclusions}
 
We have proposed LCRL, a method for guiding the training of an RL agent
using a~priori knowledge about the environment given in the form of an LTL
property, expressed via an automaton known as LDBA.  This additional knowledge, 
as we have observed in many experiments presented in this work,
improves training drastically.  We have shown that the policy synthesised
using LCRL is guaranteed to maximise the probability of satisfying the LTL
property: this architecture is shown to be working across many environments, whether with finite or continuous states and actions.  This direct relationship between the expected return of the policy
generated using LCRL and the maximum probability of satisfaction enables us
to quantify the degree of safety of the generated policy for any given
state.

Good avenues for future work are problems in which the specification is
initially unknown and has to be discovered along the learning
process~\cite{deepsynth}.  Furthermore, there are relevant synergies between
the use of LTL to guide an agent and LTL to restrict the exploration of the
agent during training (``learning while staying safe'')~\cite{cautiousRL}. 
Multi-agent setups, in which a (heterogeneous) set of agents collaborate to
satisfy an LTL formula is an interesting research direction with various
applications~\cite{hammond2021multi}.

\if\doctype1
\clearpage
\begin{appendices}
\section{Proofs}
Before proving Proposition~\ref{g_monitor_proposition}, let us present the following definition.
\begin{definition}[G-sub-formula]\label{g_monitor}
	Given an LTL property $ \varphi $ and a set of G-sub-formulae
	$\mathcal{G}$\footnote{A G-sub-formula is a sub-formula of $ \varphi $ of
		the form $ \square(\cdot) $.} we define $ \varphi[\mathcal{G}] $ the
	resulting formula when we substitute $ \mathit{true} $ for every
	G-sub-formula in $ \mathcal{G} $ and $ \neg \mathit{true} $ for other
	G-sub-formulae of~$\varphi$.
\end{definition}
\renewcommand*{\thesection}{\arabic{section}}

\setcounter{section}{3}
\setcounter{proposition}{0}

\begin{proposition}\label{g_monitor_proposition_appendix}
Given an LTL formula $ \varphi $ and its associated LDGBA
$\mathfrak{A}=\allowbreak(\mathcal{Q},q_0,\allowbreak\Sigma,\allowbreak
\mathcal{F},\allowbreak  \Delta)$, the set members of  $ \mathds{A} $ only
depend on the current state of the automaton and not on the sequence of
automaton states that have been already visited.
\end{proposition}
\noindent\textbf{\textit{proof.}}\\*
	Let $ \mathcal{G}=\{\square\zeta_1,...,\square\zeta_f\} $ be the set of all G-sub-formulae of $ \varphi $. Since elements of $ \mathcal{G} $ are sub-formulae of $ \varphi $ we can assume an ordering over $ \mathcal{G} $, so that if $ \square\zeta_i $ is a sub-formula of $ \square\zeta_j $, then $ j>i $. In particular, $ \square\zeta_f $ is not a sub-formula any G-sub-formula. The accepting component of the LDGBA $ \mathcal{Q}_D $ is a product of $ f $ of the DBAs $ \{\textbf{D}_1,....,\textbf{D}_f\} $ called G-monitors, such that each $ \textbf{D}_i = (\mathcal{Q}_i,{q_i}_0,\Sigma,F_i,\delta_i)$ expresses $ \square\zeta_i[\mathcal{G}] $, where $\mathcal{Q}_i $ is the state space of the $ i $-th G-monitor, $ \Sigma=2^{\mathcal{AP}} $, and $ \delta_i:\mathcal{Q}_i\times\Sigma\rightarrow\mathcal{Q}_i $~\cite{sickert}. Note that $ \zeta_i[\mathcal{G}] $ has no G-sub-formula any more (Definition~\ref{g_monitor}). The states of the G-monitor $ \textbf{D}_i $ are pairs of formulae where at each state the the G-monitor only checks if the run satisfies $ \square\zeta_i[\mathcal{G}] $ while putting the next G-sub-formula in the ordering of $ \mathcal{G} $ on hold, assuming that it is $ \mathit{true} $.  
	
	The product of G-monitor DBAs is a deterministic generalised B\"uchi automaton: 
	$$\textbf{P}_\textbf{D}=(\allowbreak\mathcal{Q}_D,\allowbreak {q_D}_0,\allowbreak\Sigma, \allowbreak\mathcal{F}, \allowbreak\delta)$$
	where $ \mathcal{Q}_D=\mathcal{Q}_1\times...\times\mathcal{Q}_f $, $ \Sigma=2^{\mathcal{AP}} $, $ \mathcal{F}=\{F_1,...,F_f\} $, and $ \delta=\delta_1\times...\times\delta_f $.
	
	As shown in~\cite{sickert}, while a word $ w $ is being read by the accepting component of the LDGBA, the set of G-sub-formulae that hold is ``monotonically'' expanding. If $ w \in \mathit{Words}(\varphi) $, then eventually all G-sub-formulae become $\mathit{true}$. Now, let the current state of the automaton be $ q_D=(q_1,...,q_i,...,q_f) $ while the automaton is checking whether $ \square\zeta_i[\mathcal{G}] $ is satisfied or not, assuming that $ \square\zeta_{i+1} $ is already $\mathit{true}$ (though needs to be checked later), while all G-monitors $ \square\zeta_j[\mathcal{G}],~1\leq j \leq i-1 $ have accepted $ w $. At this point, the accepting frontier set is $ \mathds{A}=\{F_{i},F_{i+1},...,F_f\} $. Reasoning by contradiction, assume that the automaton returns to $ q_D $ but $ \mathds{A}\not =\{F_{i},F_{i+1},...,F_f\} $, then at least one accepting set $ F_j,~j > i $ has been removed from $ \mathds{A} $. This means that $ \square\zeta_j $ is a sub-formula of $ \square\zeta_i $, violating the ordering of check on $ \mathcal{G} $. This is a contradiction with respect to the ordering of G-sub-formulae.

\clearpage
\begin{thm}\label{thm:1_appendix}
	\hos{Let $\varphi$ be the given LTL property and $\mathfrak{M}_\mathfrak{A}$ be the product MDP constructed by synchronising the MDP~$\mathfrak{M}$ and the LDGBA~$\mathfrak{A}$ expressing $\varphi$. There exists a discount factor that is close enough to $1$ under which an optimal Markov policy on $\mathfrak{M}_\mathfrak{A}$ that maximises the expected return over the reward in \eqref{thereward}, also maximises the probability of satisfying $\varphi$. This optimal Markov policy induces a finite-memory policy on the MDP~$\mathfrak{M}$.} 
\end{thm}

\noindent\textbf{\textit{proof.}}\\*
	Assume that there exists a policy
	$\overline{\pi}$ that satisfies $\varphi$ with maximum (non-zero) probability. Policy $\overline{\pi}$ induces a Markov chain $\mathfrak{M}_\mathfrak{A}^\mathit{\overline{{\pi}}}$ when it is applied over the MDP $\mathfrak{M}_\mathfrak{A}$. This Markov chain comprises a disjoint union between a set of transient states $\mathfrak{T}_{\overline{\pi}}$ and $h$ sets of irreducible recurrent classes $\mathfrak{R}^i_{\overline{\pi}},~i=1,...,h$~\cite{stochastic}, namely:
	$$
	\mathfrak{M}_\mathfrak{A}^\mathit{\overline{{\pi}}}=\mathfrak{T}_{\overline{\pi}} \sqcup \mathfrak{R}^1_{\overline{\pi}} \sqcup ... \sqcup \mathfrak{R}^h_{\overline{\pi}}.
	$$
	
	From (\ref{acc}), policy $\overline{\pi}$ satisfies $\varphi$ if and only if:
	\begin{equation}\label{recurrent_class}
	\exists \mathfrak{R}^i_{\overline{\pi}} ~\mbox{s.t.}~ \forall j\in\{1,...,f\},~{F}^\otimes_j \cap \mathfrak{R}^i_{\overline{\pi}} \neq \emptyset.
	\end{equation} 
	
	The recurrent classes that satisfy \eqref{recurrent_class} are called accepting. From the irreducibility of the recurrent class $\mathfrak{R}^i_{\overline{\pi}}$ we know that all the states in $\mathfrak{R}^i_{\overline{\pi}}$ communicate with each other thus, once a trace ends up in such set, all the accepting sets are going to be visited infinitely often. Therefore, from the definition of $ \mathds{A} $ and of the accepting frontier function (Definition~\ref{frontier}), the agent receives a positive reward $ r_p $ ever after it has reached an accepting recurrent class $\mathfrak{R}^i_{\overline{\pi}}$. 
	
	There are two other possibilities concerning the remaining recurrent classes that are not accepting. A non-accepting recurrent class, name it $ \mathfrak{R}^k_{\overline{\pi}} $, either 
	
	\begin{enumerate}
		\item has no intersection with any accepting set $ F_j^\otimes $, i.e. 
		$$
		\forall j \in \{1,...,f\},~ F_j^\otimes \cap \mathfrak{R}^k_{\overline{\pi}} = \emptyset;
		$$
		
		\item or has intersection with some of the accepting sets but not all of them, i.e. 
		\begin{equation}\label{eq:jay}
		\exists J \subset 2^{\{1,...,f\}}\setminus\{1,...,f\}~\mbox{s.t.}~\forall j \in J,~ F_j^\otimes \cap \mathfrak{R}^k_{\overline{\pi}} \neq \emptyset. 
		\end{equation}
		
	\end{enumerate} 
	
	In the second case, the agent is able to visit some accepting sets but not all of them. This means that in the update rule of the frontier accepting set $ \mathds{A} $ in Definition~\ref{frontier}, the case where $ (q\in F_j) \wedge (\mathds{A}=F_j) $ will never happen since there exist always at least one accepting set that has no intersection with $ \mathfrak{R}^k_{\overline{\pi}} $. Therefore, after a limited number of times, no positive reward can be obtained, and the reinitialisation of $ \mathds{A} $ in Definition~\ref{frontier} is blocked. 
	
	Recall Definition~\ref{def:expected_utility}, where the expected return for a state ${s}^\otimes \in \mathcal{S}^\otimes$ is defined as in \eqref{gamma} and \eqref{state_dep_utility}:
	$$
	{U}^{\overline{\pi}}(s^\otimes)=\mathds{E}^{\overline{\pi}} [\sum\limits_{n=0}^{\infty} \gamma(s^\otimes_n)^{N(s^\otimes_n)}~ r_n)|s^\otimes_0=s^\otimes],
	$$
	
	In both cases, from \eqref{thereward}, for any arbitrary $ r_p>0 $ (and $ r_n=0 $), there always exists a discounting coefficient $\eta$ such that the expected return of a trace reaching $ \mathfrak{R}^i_{\overline{\pi}} $ with unlimited number of successive times attaining positive reward, is higher than the expected return of any other trace. With unlimited number of obtaining positive reward for the traces entering the accepting recurrent class $ \mathfrak{R}^i_{\overline{\pi}} $, and with a state-dependent discount factor, it can be shown that the expected return is bounded and is higher than that for non-accepting traces, which have limited number of attainment of positive rewards.
	
	In the following, by contradiction, we show that any optimal policy ${\pi}^*$ which optimises the expected return will satisfy the property with maximum probability if $\eta$ is close to one. Recall from Definition~\ref{ltlprobab} that the probability of satisfying $\varphi$ under policy $\pi$ at state $s^\otimes$ is:
	$$
	\mathit{Pr}({s^\otimes..}^{\pi} \models \varphi),
	$$ 
	where ${s^\otimes..}^{\pi}$ is the collection of all paths starting from $ s^\otimes $ under policy~$\pi$. Thus, for the policy $\overline{\pi}$ we have:
	\begin{equation}\label{eq:policy_bar}
	\overline{\pi} = \arg\sup_{\pi \in \mathcal{D}} \mathit{Pr}({s^\otimes..}^{\pi} \models \varphi).
	\end{equation}
	Accordingly, the expected return for for policy $\overline{\pi}$ can be rewritten as:
	\begin{align}\label{eq:return_split}
	\begin{aligned}
	&{U}^{\overline{\pi}}(s^\otimes)=\\
	&\mathds{E}^{\overline{\pi}} \Big[\sum\limits_{n=0}^{\infty} \gamma(s^\otimes_n)^{N(s^\otimes_n)}~ r_n \Big| s^\otimes_0=s^\otimes,~\rho=L(s_0)L(s_1)...\models\varphi\Big] 
	\mathit{Pr}({s^\otimes..}^{\overline{\pi}}\models \varphi)+\\
	&\mathds{E}^{\overline{\pi}} \Big[\sum\limits_{n=0}^{\infty} \gamma(s^\otimes_n)^{N(s^\otimes_n)}~ r_n\Big|s^\otimes_0=s^\otimes,~\rho=L(s_0)L(s_1)...\not\models\varphi\Big] 
	\mathit{Pr}({s^\otimes..}^{\overline{\pi}} \not\models \varphi).
	\end{aligned}
	\end{align}
	Of course, when the policy traces satisfy the property, with unlimited number of positive reward attainment the expected return under \eqref{gamma} is 
	\begin{align}
	\begin{aligned}\label{eq:accepting_return}
	&\mathds{E}^{\overline{\pi}} \Big[\sum\limits_{n=0}^{\infty} \gamma(s^\otimes_n)^{N(s^\otimes_n)}~ r_n\Big| s^\otimes_0=s^\otimes,~\rho=L(s_0)L(s_1)...\models\varphi\Big]\times\\
	&\mathit{Pr}({s^\otimes..}^{\overline{\pi}}\models \varphi) = \dfrac{r_p}{1-\eta} \mathit{Pr}({s^\otimes..}^{\overline{\pi}}\models \varphi).
	\end{aligned}
	\end{align}
	Once the induced traces do not satisfy the property we have:
	\begin{align}
	\begin{aligned}\label{eq:non_accepting_return}
	&\mathds{E}^{\overline{\pi}} \Big[\sum\limits_{n=0}^{\infty} \gamma(s^\otimes_n)^{N(s^\otimes_n)}~ r_n\Big| s^\otimes_0=s^\otimes,~\rho=L(s_0)L(s_1)...\not\models\varphi\Big]\times\\ 
	&\mathit{Pr}({s^\otimes..}^{\overline{\pi}} \not\models \varphi) =\hspace{-6mm}\sum\limits_{\rho \in L({s^\otimes..}^{\overline{\pi}}\not\models \varphi)}\hspace{-6mm} \mathit{Pr}(\rho \in L({s^\otimes..}^{\overline{\pi}}\not\models \varphi)) \dfrac{r_p(1-\eta^{|\overline{J}|_\rho})}{1-\eta},
	\end{aligned}
	\end{align}
	where $|\overline{J}|_\rho$ is the (finite) number of times that the trace $\rho$ intersected with the accepting frontier set $\mathds{A}$ and the agent received a positive reward (see \eqref{eq:jay}). From \eqref{eq:accepting_return} and \eqref{eq:non_accepting_return} we can reformulate \eqref{eq:return_split} as follows:
	\begin{align}\label{eq:return_split_short}
	\begin{aligned}
	{U}^{\overline{\pi}}(s^\otimes)&=\dfrac{r_p}{1-\eta}\mathit{Pr}({s^\otimes..}^{\overline{\pi}} \models \varphi)+\\ 
	&\sum\limits_{\rho \in L({s^\otimes..}^{\overline{\pi}}\not\models \varphi)} \mathit{Pr}(\rho \in L({s^\otimes..}^{\overline{\pi}}\not\models \varphi)) \dfrac{r_p(1-\eta^{|\overline{J}|_\rho})}{1-\eta}.
	\end{aligned}
	\end{align}
	
	Similarly for the optimal policy $\pi^*$ we have
	\begin{align}\label{eq:return_split_short_2}
	\begin{aligned}
	{U}^{{\pi}^*}(s^\otimes)&=\dfrac{r_p}{1-\eta}\mathit{Pr}({s^\otimes..}^{{\pi}^*} \models \varphi)+\\ &\hspace{-2mm}\sum\limits_{\rho \in L({s^\otimes..}^{{\pi}^*}\not\models \varphi)} \hspace{-2mm}\mathit{Pr}(\rho \in L({s^\otimes..}^{{\pi}^*}\not\models \varphi)) \dfrac{r_p(1-\eta^{|{J}^*|_\rho})}{1-\eta},
	\end{aligned}
	\end{align}   
	where $|{J}^*|_\rho$ is the (finite) number of times that the trace $\rho$ intersected with the accepting frontier set $\mathds{A}$. We then factorise $r_p/1-\eta$ from \eqref{eq:return_split_short} and \eqref{eq:return_split_short_2}:
	\begin{align}\label{eq:return_split_short_f}
	\begin{aligned}
	&{U}^{\overline{\pi}}(s^\otimes)=\\
	&\dfrac{r_p}{1-\eta}\Big[\mathit{Pr}({s^\otimes..}^{\overline{\pi}} \models \varphi)+  
	\sum\limits_{\rho \in L({s^\otimes..}^{\overline{\pi}}\not\models \varphi)} \mathit{Pr}(\rho \in L({s^\otimes..}^{\overline{\pi}}\not\models \varphi)) (1-\eta^{|\overline{J}|_\rho})\Big].
	\end{aligned}
	\end{align}
	\begin{align}\label{eq:return_split_short_f_2}
	\begin{aligned}
	&{U}^{{\pi}^*}(s^\otimes)=\\
	&\dfrac{r_p}{1-\eta}\Big[\mathit{Pr}({s^\otimes..}^{{\pi}^*} \models \varphi)+  
	\hspace{-1.5mm}\sum\limits_{\rho \in L({s^\otimes..}^{{\pi}^*}\not\models \varphi)}\hspace{-1.5mm} \mathit{Pr}(\rho \in L({s^\otimes..}^{{\pi}^*}\not\models \varphi))(1-\eta^{|{J}^*|_\rho})\Big].
	\end{aligned}
	\end{align}   
	Now suppose that the optimal policy ${\pi}^*$ does not satisfy the property $\varphi$ with maximum probability. Given that $\overline{\pi}$ maximises the satisfaction probability, from \eqref{eq:policy_bar} we have:
	\begin{equation}\label{eq:max_probab}
	\mathit{Pr}({s^\otimes..}^{\overline{\pi}} \models \varphi) > \mathit{Pr}({s^\otimes..}^{{\pi}^*} \models \varphi).
	\end{equation}
	At the same time, it is easy to see from \eqref{eq:return_split_short_f} and \eqref{eq:return_split_short_f_2} that:
	$$
	\lim_{\eta\rightarrow 1^-}\dfrac{{U}^{\overline{\pi}}(s^\otimes)}{{U}^{{\pi}^*}(s^\otimes)} = \dfrac{\mathit{Pr}({s^\otimes..}^{\overline{\pi}} \models \varphi)}{\mathit{Pr}({s^\otimes..}^{{\pi}^*} \models \varphi)},
	$$
	and consequently from \eqref{eq:max_probab}:
	$$
	{U}^{\overline{\pi}}(s^\otimes) > {U}^{{\pi}^*}(s^\otimes)
	$$
	This is, however, in direct contrast with Definition~\ref{optimal_pol} and the optimality of the policy $\pi^*$, leading to a contradiction. This essentially means that the optimal policy $\pi^*$ maximises the probability of satisfying $\varphi$.
	%
	%

\clearpage
\begin{corollary}[Maximum Probability of Satisfaction]\label{cor:max_sat_probab_appendix}
	From Definition \ref{ltlprobab}, for a discounting factor close enough to $1$ the maximum probability of satisfaction at any state $s^\otimes$ can be determined from the LCRL value function as 
	$$
	\mathit{Pr}_{\max}(s^\otimes \models\varphi) = \dfrac{1-\eta}{r_p}~ {U}^{{\pi}^*}(s^\otimes). 
	$$
\end{corollary}
\noindent\textbf{\textit{proof.}}\\*
	The proof is a direct consequent of  \eqref{eq:return_split_short_f_2} and Theorem \ref{thm:1_appendix} results when $\eta\rightarrow 1^-$.

\clearpage
\begin{corollary}\label{thm:2_appendix} 
If no policy in the MDP~$\mathfrak{M}$ can be generated to satisfy the
property~$ \varphi $, LCRL yields in the limit a policy that is closest
(according to the previous Definition) to satisfying the given LTL formula
$\varphi$.
\end{corollary}

\noindent\textbf{\textit{proof.}}\\*
	Assume that there exists no policy in the MDP~$\mathfrak{M}$ that can satisfy the property~$\varphi$. Construct the induced Markov chain $\mathfrak{M}_\mathfrak{A}^\pi$ for any arbitrary policy ${\pi}$ and its associated set of transient states $\mathfrak{T}_{{\pi}}$ and $h$ sets of irreducible recurrent classes~$\mathfrak{R}^i_{{\pi}}$: $$\mathfrak{M}_\mathfrak{A}^\pi=\mathfrak{T}_{{\pi}} \sqcup \mathfrak{R}^1_{{\pi}} \sqcup ... \sqcup \mathfrak{R}^h_{{\pi}}.$$
	By assumption, policy ${\pi}$ cannot satisfy the property and we thus have that $$\forall \mathfrak{R}^i_{{\pi}},~\exists j \in \{1,...,f\},~ F_j^\otimes \cap \mathfrak{R}^i_{{\pi}} = \emptyset,$$ which means that there are some automaton accepting sets like $ F_j $ that cannot be visited. Therefore, after a limited number of times no positive reward is given by the reward function $ R(s^\otimes,a) $. However, the closest recurrent class to satisfying the property is the one that intersects with more distinct accepting sets. \hos{More specifically, this is the recurrent class whose runs can partially satisfy~$\varphi$. From the LDGBA construction \cite{sickert}, by examining the accepting sets that are visited in this recurrent class we can also determine which sub-formula of~$\varphi$ is satisfiable by $\pi$.}
	
	By Definition~\ref{def:expected_utility}, for any arbitrary $ r_p>0 $ (and $ r_n=0 $), the expected return at the initial state for a trace with highest number of intersections with distinct accepting sets is maximum among other traces. Hence, an optimal policy produced by LCRL converges to a policy whose recurrent classes of its induced Markov chain have the highest number of intersections with the accepting sets of the automaton. 

\clearpage
\renewcommand*{\thesection}{\Alph{section}}
\setcounter{section}{1}
\section{Partial Satisfaction}\label{appndx:partial_sat}
\begin{definition}[Closeness to Satisfaction]
	Assume that the probability of satisfying the property~$\varphi$ for two policies $ \pi_1 $ and $ \pi_2 $ is zero. Accordingly, there are accepting sets in the automaton that have no intersection with runs of induced Markov chains $ \mathfrak{M}^{\pi_1} $ and $ \mathfrak{M}^{\pi_2} $. We say that $ \pi_1 $ is closer to satisfying the property if runs of $ \mathfrak{M}^{\pi_1} $ cross a larger number of distinct accepting sets of the automaton, than runs of $ \mathfrak{M}^{\pi_2} $. 
\end{definition}

\hos{In the following we show that even when it is infeasible to satisfy the given LTL specification, LCRL is able to find a policy whose traces are closest to satisfying~$ \varphi $. Specifically, depending on whether some accepting sets can be visited infinitely often, the LTL property can be partially satisfied, i.e. a sub-formula of~$ \varphi $.}

\begin{corollary}\label{thm:2} 
If no policy in the MDP~$\mathfrak{M}$ can be generated to satisfy the
property~$ \varphi $, LCRL yields in the limit a policy that is closest
(according to the previous Definition) to satisfying the given LTL formula
$\varphi$.
\end{corollary} 
\noindent\textbf{\textit{proof.}}\\*
	Assume that there exists no policy in the MDP~$\mathfrak{M}$ that can satisfy the property~$\varphi$. Construct the induced Markov chain $\mathfrak{M}_\mathfrak{A}^\pi$ for any arbitrary policy ${\pi}$ and its associated set of transient states $\mathfrak{T}_{{\pi}}$ and $h$ sets of irreducible recurrent classes~$\mathfrak{R}^i_{{\pi}}$: $$\mathfrak{M}_\mathfrak{A}^\pi=\mathfrak{T}_{{\pi}} \sqcup \mathfrak{R}^1_{{\pi}} \sqcup ... \sqcup \mathfrak{R}^h_{{\pi}}.$$
	By assumption, policy ${\pi}$ cannot satisfy the property and we thus have that $$\forall \mathfrak{R}^i_{{\pi}},~\exists j \in \{1,...,f\},~ F_j^\otimes \cap \mathfrak{R}^i_{{\pi}} = \emptyset,$$ which means that there are some automaton accepting sets like $ F_j $ that cannot be visited. Therefore, after a limited number of times no positive reward is given by the reward function $ R(s^\otimes,a) $. However, the closest recurrent class to satisfying the property is the one that intersects with more distinct accepting sets. \hos{More specifically, this is the recurrent class whose runs can partially satisfy~$\varphi$. From the LDGBA construction \cite{sickert}, by examining the accepting sets that are visited in this recurrent class we can also determine which sub-formula of~$\varphi$ is satisfiable by $\pi$.}
	
	By Definition~\ref{def:expected_utility}, for any arbitrary $ r_p>0 $ (and $ r_n=0 $), the expected return at the initial state for a trace with highest number of intersections with distinct accepting sets is maximum among other traces. Hence, an optimal policy produced by LCRL converges to a policy whose recurrent classes of its induced Markov chain have the highest number of intersections with the accepting sets of the automaton. 

\clearpage
\section{\rbtl{Comparison with a DRA-based Learning Algorithm}}

\rbtl{The problem of policy synthesis with RL under LTL constraints is investigated in several previous works, where the general recipe is to translate the LTL property into a DRA and then to construct a product MDP. For the sake of comparison, let us consider the example in \cite{dorsa}: a $5\times 5$ grid world where the starting state is $(0,3)$and the agent has to visit two regions infinitely often (areas $A$ and $B$ in Fig.~\ref{dorsaandus}). The agent has to also avoid the area~$C$. This property can be encoded as the following LTL formula:} 

\rbtl{\begin{equation}\label{dorsa's ltl}
	\square\lozenge A \wedge \square\lozenge B \wedge \square \neg C.
\end{equation}} 

\begin{figure}[!hb]\centering
	\scalebox{.8}{
		\begin{tikzpicture}[shorten >=1pt,node distance=2.8cm,on grid,auto] 
			\node[state,initial,accepting] (q_0)   {$q_0$}; 
			\node[state] (q_1) [above right=of q_0] {$q_1$}; 
			\node[state] (q_2) [below right=of q_1] {$q_2$}; 
			\path[->] 
			(q_0) edge  node {$\neg B \wedge \neg C$} (q_1)
			(q_0) edge node {$B \wedge \neg A \wedge \neg C$} (q_2)
			(q_2) edge [bend left=45] node {$A \wedge \neg C$} (q_0)
			(q_1) edge  node {$B \wedge \neg A \wedge \neg C$} (q_2)
			(q_2) edge [loop right] node {$\neg A \wedge \neg C$} ()
			(q_1) edge [loop above] node {$\neg B \wedge \neg C$} ();    
	\end{tikzpicture}}
	\caption{\rbtl{LDGBA expressing the LTL formula in (\ref{dorsa's ltl}) with removed transitions labelled $A \wedge B$ (since it is impossible to be at $A$ and $B$ at the same time).}}
	\label{dorsa3} 
\end{figure}

\begin{figure}[!hb] \centering \includegraphics[width=0.9\columnwidth]{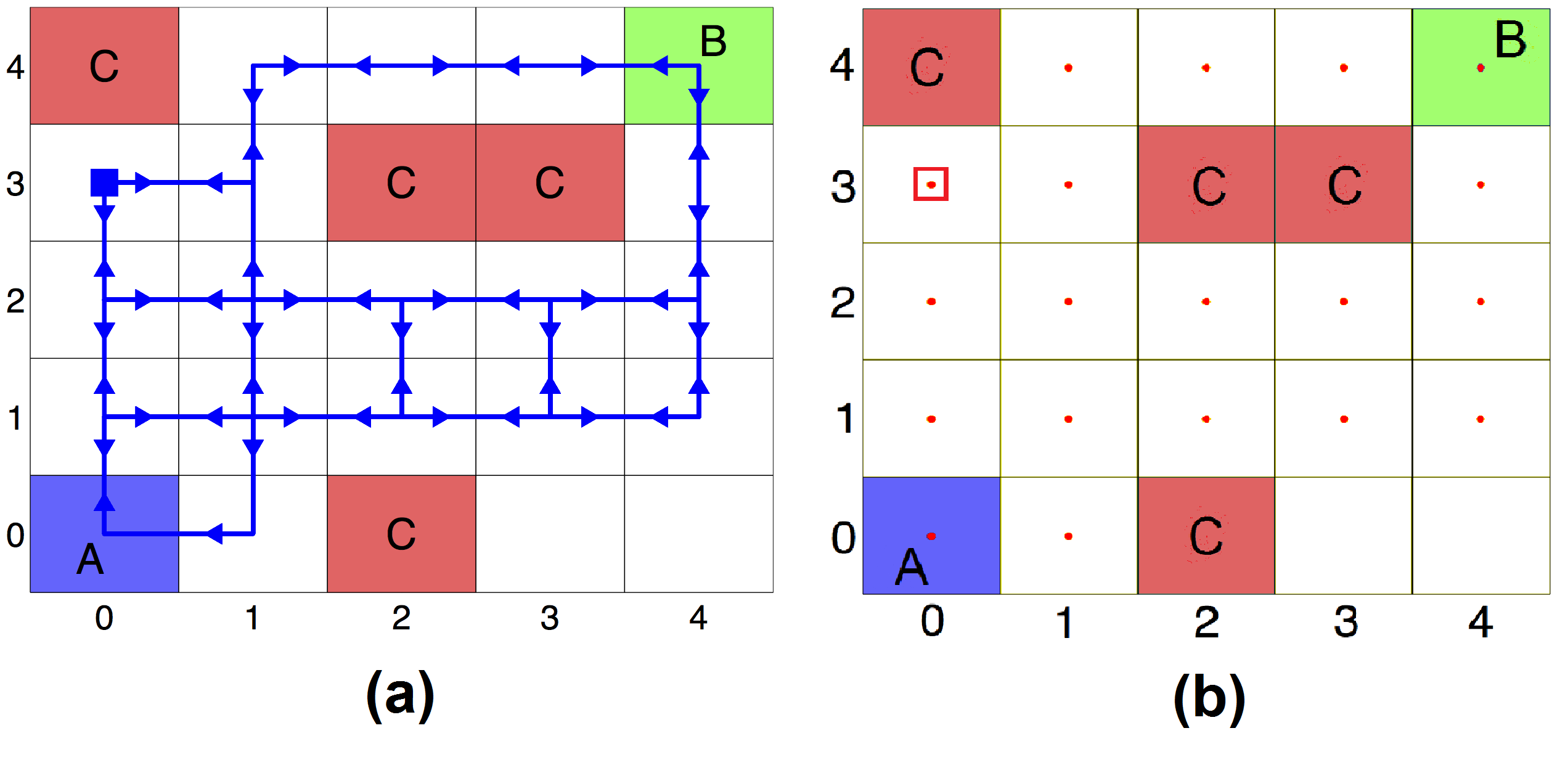} \caption{\rbtl{(a) Example considered in~\cite{dorsa}. (b) Trajectories under the policy generated by LCRL in~\cite{dorsa}.}}
	\label{dorsaandus} 
\end{figure}

\rbtl{The product MDP in~\cite{dorsa} contains 150 states, which means that the Rabin automaton has 6 states. Fig.~\ref{dorsaandus}.a shows the trajectories under the optimal policy generated by~\cite{dorsa} algorithm after 600 iterations. However, by employing LCRL we are able to generate the same trajectories with only 50 iterations (Fig.~\ref{dorsaandus}.b). The automaton that we consider is an LDGBA with only 3 states as in Fig.~\ref{dorsa3}. This result in a smaller product MDP and a much more succinct state space (only 75 states) for the algorithm to learn, which consequently leads to a faster convergence.} 

\rbtl{In addition, the reward shaping in LCRL is significantly simpler thanks to the B\"uchi acceptance condition. In a DRA $\textbf{R}(\mathcal{Q},\mathcal{Q}_0,\Sigma, \mathcal{F}, \Delta)$, the set $\mathcal{F}= \{(G_1,B_1),\ldots,\allowbreak (G_{n_F},\allowbreak B_{n_F})\}$ represents the acceptance condition in which $\allowbreak G_i,\allowbreak B_i \in \mathcal{Q}$ for $i=1,\ldots,\allowbreak n_F$. An infinite run $\theta \in \mathcal{Q}^\omega$ starting from $\mathcal{Q}_0$ is accepting if there exists $i\in \{1,\ldots,n_F\}$ such that
$$
\mathit{inf}(\theta) \cap {G}_i \neq  \emptyset \quad\mbox{and}\quad \mathit{inf}(\theta) \cap {B}_i =  \emptyset.
$$} 

\rbtl{Therefore for each $i\in \{1,\ldots,\allowbreak n_F\}$ a separate reward assignment is
needed in~\cite{dorsa} which complicates the implementation and increases the required calculation costs. This complicated reward assignment is not needed by employing the accepting frontier function in our scheme.}

\rbtl{More importantly, LCRL is a model-free learning algorithm that does not require an approximation of the transition probabilities of the underlying MDP. This even makes LCRL more easier to employ. We would like to emphasize that LCRL convergence proof solely depends on the structure of the MDP and this allows LCRL to find satisfying policies even if they have probability of less than one.}

\clearpage
\section{Alternatives to LCNFQ}
In the following, we discuss the most popular alternative approaches to solving infinite-state MDPs, namely the Voronoi Quantiser (VQ) and Fitted Value Iteration (FVI). 

\subsection{Voronoi Quantiser}\label{vorosection}

VQ can be classified as a discretisation algorithm which abstracts the continuous-state MDP to a finite-state MDP, allowing classical RL to be run. However, most of discretisation techniques are usually done in an ad-hoc manner, disregarding one of the most appealing features of RL: autonomy. In other words, RL is able to produce the optimal policy with regards to the reward function, with minimum supervision. Therefore, the state space discretisation should be performed as part of the learning task, instead of being fixed at the start of the learning process.

\begin{algorithm2e}[!b]
	\DontPrintSemicolon
	\SetKw{return}{return}
	\SetKwRepeat{Do}{do}{while}
	\SetKwData{conflict}{conflict}
	\SetKwData{safe}{safe}
	\SetKwData{sat}{sat}
	\SetKwData{unsafe}{unsafe}
	\SetKwData{unknown}{unknown}
	\SetKwData{true}{true}
	\SetKwInOut{Input}{input}
	\SetKwInOut{Output}{output}
	\SetKwFor{Loop}{Loop}{}{}
	\SetKw{KwNot}{not}
	\begin{small}
		\Input{minimum resolution $\Delta$}
		\Output{approximated Q-function $Q$}
		initialize $c_1=c=$ initial state\;
		initialize $Q(c_1,a)=0,~\forall a \in \mathcal{A}$\;
		\Repeat{end of trial}
		{
			set $\mathcal{C}=c_1$ \hfill\;
			$a_*=\arg\max_{a\in\mathcal{A}} Q(c,a)$\;
			\Repeat{end of trial}
			{
				execute action $a_*$ and observe the next state $(s',q)$\;
				\uIf{$\mathcal{C}^{q}$ is empty}
				{
					append $c_{\mathit{new}}=(s',q)$ to $\mathcal{C}^{q}$\;
					initialize $Q(c_{\mathit{new}},a)=0,~\forall a \in \mathcal{A}$\;
				}
				\Else{
					determine the nearest neighbour $c_{\mathit{new}}$ within $\mathcal{C}^q$\;
					\uIf{$c_{\mathit{new}}=c$}
					{
						
						\If{$||c-(s',q)||_2>\Delta$}
						{
							append $c_{\mathit{new}}=(s',q)$ to $\mathcal{C}^{q}$\;
							initialize $Q(c_{\mathit{new}},a)=0,~\forall a \in \mathcal{A}$\;
						}
						
					}
					\Else{
						
						$Q(c,a_*)=(1-\mu)Q(c,a_*)+\mu
						[R(c,a_*)+\gamma \max\limits_{a'}(Q(c_{\mathit{new}},a'))]$
					}
				}
				$c=c_{\mathit{new}}$\;
			}
		}
	\end{small}
	\caption{Episodic VQ}
	\label{voronoial}
\end{algorithm2e}

Inspired by~\cite{voronoi}, we propose a version of VQ that is able to discretise the state space of the product MDP $\mathcal{S}^\otimes$, while allowing RL to explore the MDP. VQ maps the state space onto a finite set of disjoint regions called Voronoi cells~\cite{voronoi_o}. The set of centroids of these cells is denoted by $\mathcal{C}=\{c_i\}_{i=1}^m,~c_i \in  \mathcal{S}^\otimes$, where $m$ is the number of the cells. With $ \mathcal{C} $, we are able to use QL and find an approximation of the optimal policy for a continuous-state MDP. 

In the beginning, $\mathcal{C}$ is initialised to consist of just one $c_1$, which corresponds to the initial state. This means that the agent views the entire state space as a homogeneous region when no a-priori knowledge is available. Assuming that states are represented by vectors, when the agent explores this unknown state space, the Euclidean norm of the distance between each newly visited state and its nearest neighbour can calculated. If this norm is greater than a threshold value $\Delta$ called ``minimum resolution'', or if the new state $ s^\otimes $ comprises an automaton state that has never been visited, then the newly visited state is appended to $\mathcal{C}$. Therefore, as the agent continues to explore, the size of $\mathcal{C}$ would increase until the ``relevant'' parts of the state space are partitioned. In our algorithm, the set $\mathcal{C}$ has $|\mathcal{Q}|$ disjoint subsets where $\mathcal{Q}$ is the finite set of states of the automaton. Each subset $\mathcal{C}^{q_j},~j=1,...,|\mathcal{Q}|$ contains the centroids of those Voronoi cells that have the form of $c_i^{q_j}=(\cdot,q_j)$, i.e. $ \bigcup_i^m c_i^{q_j} = \mathcal{C}^{q_j}$ and $ \mathcal{C}=\bigcup_{j=1}^{|\mathcal{Q}|} \mathcal{C}^{q_j}$. Therefore, a Voronoi cell
$$
\{(s,q_j) \in \mathcal{S}^\otimes,||(s,q_j)-c_i^{q_j}||_2\leq||(s,q_j)-c_{i'}^{q_j}||_2 \},
$$
is defined by the nearest
neighbour rule for any $i'\neq i$. The proposed VQ algorithm is presented in Algorithm~\ref{voronoial}.

\subsection{Fitted Value Iteration}\label{FVIsection}
FVI is an approximate DP algorithm for continuous-state MDPs, which employs function approximation techniques~\cite{gordon}. 
In standard DP the goal is to find a mapping, i.e. value function, from the state space to $\mathbb{R}$, which can generate the optimal policy. The value function in our setup is the expected reward in (\ref{upol}) when $\pi$ is the optimal policy, i.e. ${U}^{\pi^*}$. Over continuous state spaces, analytical representations of the value function are in general not available. Approximations can be obtained numerically through approximate value iterations, which involve approximately iterating a Bellman operator on a an approximate value function~\cite{cdp}. 

We propose a modified version of FVI that can handle the product MDP. The global value function $v:\mathcal{S}^\otimes\rightarrow\mathbb{R}$, or more specifically $v:\mathcal{S}\times\mathcal{Q}\rightarrow\mathbb{R}$, consists of $|\mathcal{Q}|$ number of components. For each $ q_j\in\mathcal{Q} $, the sub-value function $v^{q_j}:\mathcal{S}\rightarrow\mathbb{R}$ returns the value the states of the form $(s,q_j)$. Similar to the LCNFQ algorithm, the components are not decoupled. 

Let $P^\otimes(dy|s^\otimes,a)$ be the distribution over $\mathcal{S}^\otimes$ for the successive state given that the current state is $s^\otimes$ and the selected action is $a$. For each state $ (s,q_j) $, the Bellman update over each component of value function $v^{q_j}$ is defined as:

\begin{equation}\label{bellman}
\tau v^{q_j}(s)=\sup\limits_{a\in\mathcal{A}} \{\int v(y) P^\otimes(dy|(s,q_j),a)\},
\end{equation}
where $\tau$ is the Bellman operator~\cite{hernandez}. 
The update in \eqref{bellman} is different than the standard Bellman update in DP, as it does not comprise a running reward, and as the (terminal) reward is replaced by the following function initialization: 
\begin{equation}
\label{fviinit}
v(s^\otimes) = \left\{
\begin{array}{ll}
r_p & $ if $ {s^\otimes} \in \mathbb{A},\\
r_n & $ otherwise. 
$
\end{array}
\right.
\end{equation}


The main hurdle in executing the Bellman operator in continuous state MDPs, as in (\ref{bellman}), is that no analytical representations of the value function $v$ and of its components $ v^{q_j},~q_j\in\mathcal{Q} $ are in general available. 
Therefore, we employ an approximation method, by introducing a new operator $L$.

The operator $L$ provides an approximation of the value function, denoted by $ Lv $, and of its components $v^{q_j}$, which we denote by $Lv^{q_j}$. For each $ q_j\in\mathcal{Q} $ the approximation is based on a set of points $\{(s_i,q_j)\}_{i=1}^k \subset \mathcal{S}^\otimes$ which are called centres. For each $ q_j $, the centres $ i=1,...,k $ are distributed uniformly over $ \mathcal{S} $. 

In the proposed FVI algorithm, we employ the kernel averager method~\cite{cdp}, which can be represented by the following expression for each state $ (s,q_j) $:
\begin{equation}\label{kernel}
Lv(s,q_j)=Lv^{q_j}(s)=\dfrac{\sum_{i=1}^{k} K(s_i-s) v^{q_j}(s_i)}{\sum_{i=1}^{k} K(s_i-s)},
\end{equation}
where the kernel $K:\mathcal{S}\rightarrow\mathbb{R}$ is a radial basis function, such as $e^{-|s-s_i|/{h'}}$, and $h'$ is smoothing parameter. Each kernel is characterised by the point $s_i$, and its value decays to zero as $s$ diverges from $s_i$. This means that for each $ q_j\in\mathcal{Q} $ the approximation operator $L$ in (\ref{kernel}) is a convex combination of the values of the centres $\{s_i\}_{i=1}^{k}$ with larger weight given to those values $v^{q_j}(s_i)$ for which $s_i$ is close to $s$. Note that the smoothing parameter $h'$ controls the weight assigned to more distant values. 

\begin{algorithm2e}[!t]
	\DontPrintSemicolon
	\SetKw{return}{return}
	\SetKwRepeat{Do}{do}{while}
	\SetKwData{conflict}{conflict}
	\SetKwData{safe}{safe}
	\SetKwData{sat}{sat}
	\SetKwData{unsafe}{unsafe}
	\SetKwData{unknown}{unknown}
	\SetKwData{true}{true}
	\SetKwInOut{Input}{input}
	\SetKwInOut{Output}{output}
	\SetKwFor{Loop}{Loop}{}{}
	\SetKw{KwNot}{not}
	\begin{small}
		\Input{MDP $\mathfrak{M}$, a set of samples $\{s^\otimes_i\}_{i=1}^k=\{(s_i,q_j)\}_{i=1}^k$ for each $q_j\in\mathcal{Q}$, Monte Carlo sampling number $Z$, smoothing parameter $h'$}
		\Output{approximated value function $Lv$}
		initialize $Lv$ \;
		sample $\mathcal{Y}_a^Z(s_i,q_j),~\forall q_j \in \mathcal{Q},~\forall i=1,...,k~,~\forall a\in\mathcal{A}$\;
		\Repeat{end of trial}
		{
			\For{$j=|\mathcal{Q}|$ \textbf{to} $1$}
			{
				$~\forall q_j \in \mathcal{Q},~\forall i=1,...,k~,~\forall a\in\mathcal{A}$ calculate $I_a((s_i,q_j))=1/Z \sum_{y \in \mathcal{Y}_a^Z(s_i,q_j)} Lv(y)$ using \eqref{kernel}\;
				for each state $(s_i,q_j)$, update $v^{q_j}(s_i)=\sup_{a\in\mathcal{A}}\{I_a((s_i,q_j))\}$ in  \eqref{kernel}
			}
		}
	\end{small}
	\caption{FVI}
	\label{FVI}
\end{algorithm2e}

In order to approximate the integral in the Bellman update (\ref{bellman}) we use a Monte Carlo sampling technique~\cite{montec}. For each centre $(s_i,q_j)$ and for each action $a$, we sample the next state $y_a^z(s_i,q_j)$ for $z=1,...,Z$ times and append these samples to the set of $ Z $ subsequent states $\mathcal{Y}_a^Z(s_i,q_j)$. We then replace the integral with 
\begin{equation}\label{montec}
I_a(s_i,q_j)=\dfrac{1}{Z} \sum\limits_{z=1}^{Z} Lv(y_a^z(s_i,q_j)).
\end{equation}

The approximate value function $ Lv $ is initialised according to
\eqref{fviinit}.  In each loop in FVI, the Bellman update approximation is
first executed over those sub-value functions that are linked with the
accepting states of the LDGBA, i.e.  those that have an initial value of
$r_p$.  The approximate Bellman update then goes backward until it reaches
those sub-value functions that are linked with the initial states of the
automaton.  This allows the state values to back-propagate through the
product MDP transitions that connects the sub-value function via
\eqref{montec}.  Without loss of generality we assume that the automaton
states are ordered and hence the back-propagation starts from
$q_i=|\mathcal{Q}|$.  Once we have the approximated value function, we can
generate the optimal policy by following the maximum value
(Algorithm~\ref{FVI}).

We conclude this section by emphasising that Algorithms~\ref{voronoial} and~\ref{FVI} are proposed to be benchmarked against LCNFQ. Further, MDP abstraction techniques such as~\cite{faust} failed to scale and to find an optimal policy.

\end{appendices}
\fi
\if\doctype3

\fi
\clearpage
\bibliographystyle{elsarticle-num}
\bibliography{Biblio}           

\end{document}